\RequirePackage{etex}

\documentclass{article}

\usepackage{microtype}
\usepackage{graphicx}
\usepackage{subfigure}
\usepackage{booktabs} 

\usepackage{etex}

\usepackage{hyperref}

\usepackage[accepted]{icml2024}

\usepackage{amsmath}
\usepackage{amssymb}
\usepackage{mathtools}
\usepackage{amsthm}

\usepackage{microtype}
\usepackage{graphicx}
\usepackage{subcaption}
\usepackage{lipsum}
\usepackage{array} 
\usepackage{etex}
\usepackage{adjustbox}

\newcommand\sm{$S^{*}_{m}$}

\usepackage[capitalize,noabbrev]{cleveref}

\theoremstyle{plain}

\theoremstyle{definition}

\theoremstyle{remark}

\usepackage[textsize=tiny]{todonotes}

\icmltitlerunning{Not Every Image is Worth a Thousand Words: Quantifying Originality in Stable Diffusion}

\begin{document}

\twocolumn[

\icmltitle{Not Every Image is Worth a Thousand Words:  \\ Quantifying Originality  in Stable Diffusion}




\icmlsetsymbol{equal}{*}

\begin{icmlauthorlist}
\icmlauthor{Adi Haviv}{yyy}
\icmlauthor{Shahar Sarfaty}{yyy}
\icmlauthor{Uri Hacohen}{sch}
\icmlauthor{Niva Elkin-Koren}{sch}
\icmlauthor{Roi Livni}{comp}
\icmlauthor{Amit H Bermano}{yyy}
\end{icmlauthorlist}

\icmlaffiliation{yyy}{School of Computer Science, Tel Aviv University}
\icmlaffiliation{sch}{Faculty of Law, Tel Aviv University}
\icmlaffiliation{comp}{School of Electrical Engineering, Tel Aviv University}

\icmlcorrespondingauthor{Adi Haviv}{adi.haviv@cs.tau.ac.il}

\icmlkeywords{Machine Learning, ICML}

\vskip 0.3in
]



\printAffiliationsAndNotice{\icmlEqualContribution} 

\begin{abstract}
This work addresses the challenge of quantifying originality in text-to-image (T2I) generative diffusion models, with a focus on copyright originality. We begin by evaluating T2I models' ability to innovate and generalize through controlled experiments, revealing that stable diffusion models can effectively recreate unseen elements with sufficiently diverse training data. Then, our key insight is that concepts and combinations of image elements the model is familiar with, and saw more during training, are more concisly represented in the model's latent space. We hence propose a method that leverages textual inversion to measure the originality of an image based on the number of tokens required for its reconstruction by the model. Our approach is inspired by legal definitions of originality and aims to assess whether a model can produce original content without relying on specific prompts or having the training data of the model. We demonstrate our method using both a pre-trained stable diffusion model and a synthetic dataset, showing a correlation between the number of tokens and image originality. This work contributes to the understanding of originality in generative models and has implications for copyright infringement cases.
\end{abstract}

\section{Introduction}
\label{sec:intro}

\begin{figure}[t]
    \centering
    \includegraphics[width=.92\linewidth]{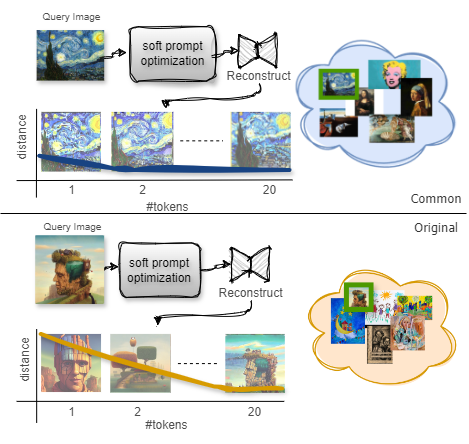} 
    \vspace{-0.6cm}
    \caption{Illustration of our approach for measuring image originality using multi-token textual inversion. Original images require more tokens for accurate reconstruction, while common images like Van Gogh's "Starry Night" need only one token.}
    \vspace{-0.6cm}
    \label{fig:main}
\end{figure}

Large-scale Text-to-Image (T2I) Generative Diffusion-based models have revolutionized our ability to generate and manufacture visual content using natural language descriptions. T2I models, as their name suggests, are designed to produce images given a textual prompt. Distinctively from a search engine, T2I are not meant to find an existing image that fits a certain description, but they are supposed to \emph{generate} novel content that fits the description of the text. Despite their aforementioned design purpose quantifying originality remains a formidable challenge both in practice as well as in theory.  

This challenge is not solely scholastic and arises in the context of legal concerns surrounding copyright laws, where T2I models, trained on expansive datasets like LAION-5B \cite{schuhmann2022laion} that include copyrighted materials, are often at the center of infringement accusations. Here too, quantifying originality poses a challenge as copyright law 
only protects the aspects of expressive works deemed \emph{original} by the judiciary \cite{harper1985,feist1991,uscode1990}, where originality necessitates a minimal degree of creativity and authorship \cite{feist1991}.

In turn, methodologically sound methods for demonstrating creativity and originality in a T2I model become a pressing matter. Traditional strategies often formalize the problem of not-copying as a form of memorization constraint that inhibits overfitting of the data \cite{carlini2023extracting, bousquet2020synthetic, vyas2023provable}. This is also highlighted in the recently implemented EU AI Act, which mandates the disclosure of training data \cite{IViR2023} that requires greater transparency in the operation and training of these models. However, regulating memorization is not necessarily aligned with the purpose of copyright law \cite{elkin2023can},  can be overly restrictive, and also poses computational as well as statistical challenges \cite{feldman2020does, feldman2020neural, attias2024information, livni2024information, zhang2016understanding}.

In this paper we consider an alternative viewpoint. Instead of looking at the training data and what information it holds, we analyze the model itself and what it had actually learned from the information the data has to offer. Specifically, We investigate whether T2I models can, in fact themselves, be utilized to discriminate between generic and original content, according to their understanding of the world. 

Towards this goal, we start with a set of prerequisite, controlled experiments on synthetic data that establish T2I models' ability to generalize. We then move on and propose a quantitative framework that assesses the originality of images based on the model's familiarity with the training data. Finally, we implement our framework concretely and provide a set of synthetic as well as real-world data experiments, that demonstrate the potential of T2I models in identifying originality in output content. We believe that our framework can be harnessed to build further metrics for originality and genericity, which in turn can be used to audit the utility of generative models, and hopefully be used to analyze originality in real-world as well as synthetic images.

We begin by assessing how well T2I models can innovate and generalize in a controlled set of experiments. Experiments to assess the generalization of generative models \cite{zhao2018bias} have been conducted in previous work, but the effects of textual conditioning have yet to be explored. Moreover, we show how we can exploit textual conditioning for a series of new experiments that deepen our understanding of generalization. 

Our experiments reveal that stable diffusion models are particularly adept at adapting to and recreating unseen elements when sufficiently diversified prompts are used.
Overall, our findings underscore the critical importance of training models on diverse and comprehensive datasets \cite{lemley2020fair}.

Next, we introduce our conceptual framework to quantitatively measure originality or genericity of images, followed by a practical implementation of it. Inspired by the theoretical work of \citet{scheffler2022formalizing}, we look at the complexity of description as a measure of originality. The working hypothesis is that common concepts are easier to describe in the machine's language (i.e., the latent space) than original concepts. Unlike \citet{scheffler2022formalizing} that builds on the notion of Kolmogorov complexity, 
Similarly, we observe that the latent representation's length is also a great evaluator for complexity, since generic concepts, that have been seen many times during training. We accordingly search for the shortest latent representation, using recent literature \cite{gal2022image}. By applying textual inversion techniques, we evaluate the extent to which a concept is familiar to the model, and thus, potentially unoriginal.

Finally, we validate our framework with empirical experiments utilizing both a widely used pre-trained stable diffusion model and a custom-trained model designed specifically for this study, which processes synthetic data composed of various shapes, colors, sizes, and infills. These experiments employ both textual inversion and the DreamSim algorithm to analyze the correlation between the ease of concept recreation—measured by the number of tokens needed—and the originality of the images relative to the training dataset \cite{fu2023dreamsim}.
Our experiments reaffirm and validate that embracing rather than avoiding memorization might enable generative models to produce more innovative and diverse content.  

Overall we contribute to the study of originality and copyright in generative models by suggesting a new technique to identify genericity, as well as offering a methodology of synthetic and real experiments that can further be advanced in order to assess and improve generative models.

\section{T2I Models Produce Original Content}
\label{sec:analysis}

\begin{figure*}[t]
    \centering
    \includegraphics[width=.90\linewidth]{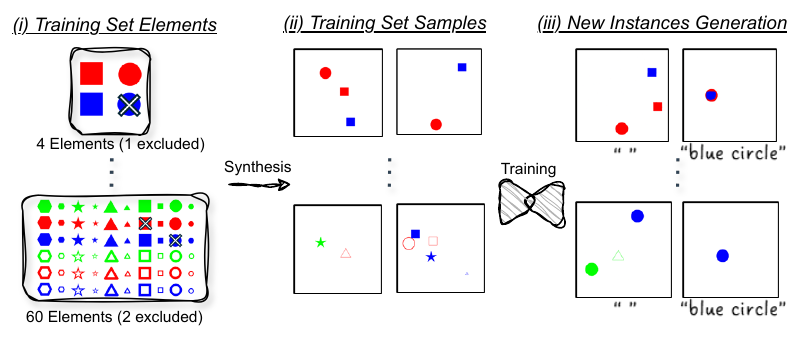} 
    \caption{Generalization experiments diagram on synthetic data. \textbf{(i)} We evaluate the relationship between data diversity and originality by running experiments over sets of distinct elements in increasing sizes. \textbf{(ii)} Examples of datasets synthesized from the respective element sets illustrate the variety within the data. \textbf{(iii)} T2I models trained from scratch using the corresponding datasets, with images generated by prompting the models with either an empty prompt or specific element descriptions.}
    
    \label{fig:synthetic_gen_diag}
    
\end{figure*}

Before presenting our general framework, we first conduct preliminary experiments to establish T2I models ability to generalize and generate original content. Such experiments are prerequisite to any attempt to quantify such originality. The generalization abilities of generative models have been explored prior to the rise in T2I models popularity \cite{zhao2018bias}. The effect, though, of textual conditioning via prompts on generalization remains unexplored. In this section we demonstrate, using the synthetic setup, that while diffusion T2I models obviously can memorize details from the training data and generate copied versions of it in the output, they can also generalize to new concepts and content, through composition of seen properties, surprisingly well. We then investigate the extent to which these models can generalize, contingent upon the distribution of training data and prompt guidance.

\begin{figure*}[t]
    \centering
    \vspace{-0.1cm}
    \includegraphics[width=.99\linewidth]{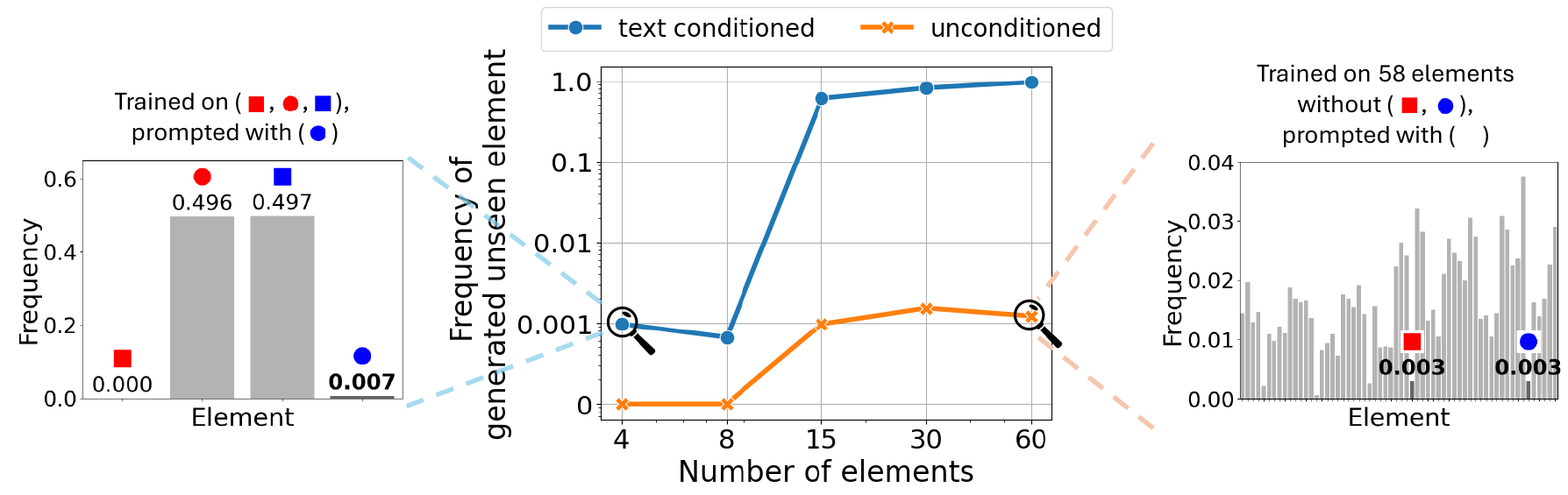} 
    
    \vspace{-0.2cm}
    \caption{Synthetic generalization experiments results. \textbf{Center:} Generalization capability of the trained models vs. training data diversity (x-axis) and conditioning types (blue line vs. orange line). \textbf{Sides:} Detailed distributions for a specific set and missing elements. These results support the notion that models generate both original and reproduced content, highly depending on the training data.}
    \label{fig:analysis_results}
    \vspace{-0.4cm}
\end{figure*}

We introduce a generalization assessment setup, as depicted in \cref{fig:synthetic_gen_diag}, and present experimental findings in Fig.~\ref{fig:analysis_results} and expend on those in~\cref{sec:appx_syn_framework}. Our quantitative analysis reveals that generalization improves with increased training data diversity and textual conditioning. Additionally, we observe an enhancement in the quality of generated images with greater training data diversity.

\paragraph{Setup and Methodology}
The experiment evaluates the model's ability to generalize by withholding specific elements during training and assessing their generation post-training. Each element in the dataset has four dimensions: Size, Color, Texture, and Shape Type. The dataset's diversity ranges from minimal, with two shape types (square and circle) and two colors (red and blue), to maximal, with five shape types (square, circle, triangle, hexagon, and star), three colors (red, green, and blue), two sizes (big and small), and two textures (full and empty), creating 60 unique elements. The degree of generalization is quantified by the frequency of the missing element's occurrence in the generated set. 

This experiment is conducted twice: using an empty prompt and with a prompt describing the missing element. Results are averaged over multiple experiments with different spanning sets and missing elements. Additional details on the synthetic framework setup and methodology is provided in~\cref{sec:appx_syn_framework} and an illustration of the experiment is provided in~\cref{fig:synthetic_gen_diag}.

\paragraph{Data Diversity Promotes Generalizability}
The synthetic experiments yield evidence that, indeed, the models are capable of generalizing and generating novel content. Results are summarized in Fig.~\ref{fig:analysis_results}, which depicts how diversity in the training data enhances generalization. Prompting allows us to further exemplify this by actively requesting new unobserved content.

Our results show that increasing training data diversity helps the model to generalize. By prompting a request to an element not seen during training, we can see that a dataset containing as few as 60 unique elements yields the requested element consistently. This demonstrates the model's ability to deconstruct and reconstruct elements and effectively translate them between textual and visual domains. When the model is trained on monotonous datasets, namely datasets with relatively few elements, the model collapses into the behavior of copying and fails to reconstruct novel elements. In a typical example, in the text-conditioned simple model, a model that was trained on blue squares and red circles was prompted with "blue circle". The model generated images with two elements, a blue square \emph{and} red circle, but failed to generate the novel concept of blue circle. Interestingly, the quality and expressiveness seem to correlate. This is illustrated in~\Cref{fig:synthetic_gen_diag}(iii). 

Quality improves with training data diversity alongside the generalization frequency. For example, comparing the generated blue circles of the simple model with those of the diverse model, the diverse model consistently produces higher-quality results, regardless of conditioning type. This trend is consistent across various elements, as demonstrated in the supplementary. These observations highlight the significance of dataset composition in fostering model creativity and robustness.

Overall, these experiments establish that the model does not just memorize the data but can compose unseen elements and concepts; we can, therefore, now proceed to measure originality with such models.

\section{Measuring Originally Using Conditioned Text}
\label{sec:method}

Before presenting our method, we provide a background overview for some key-ingredient methods that we use within our framework, Stable diffusion \cite{rombach2022high} as our T2I models architecture and a variant of textual inversion method \cite{gal2022image} as the process for measuring complexity. 


\begin{figure*}[t]
    \centering
    \includegraphics[width=0.98\linewidth]{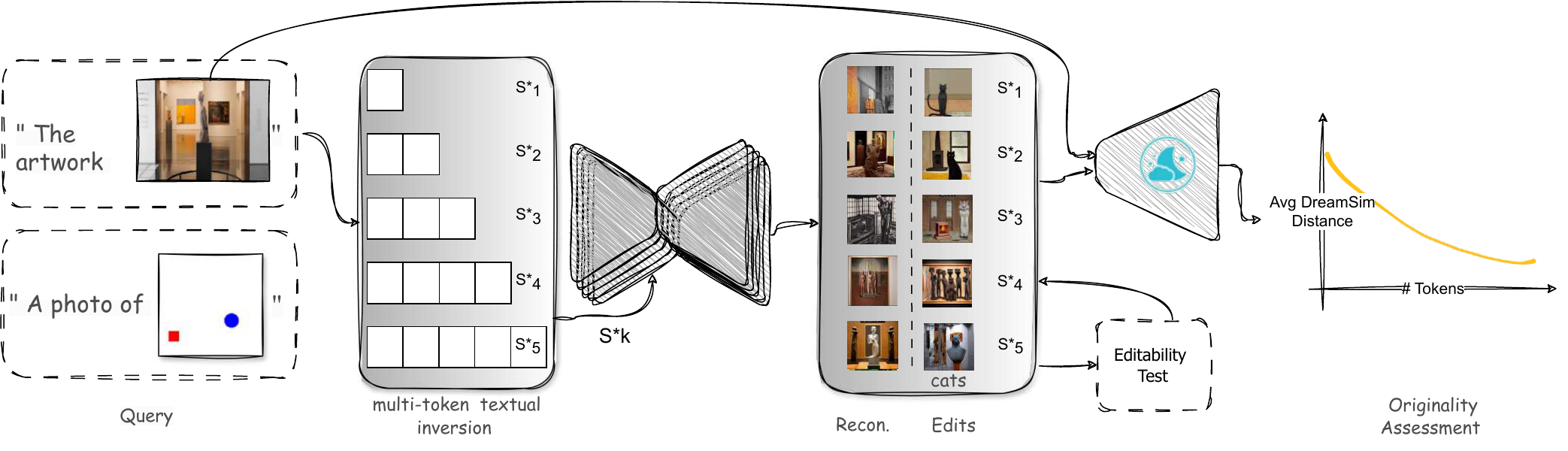} 
    \caption{Method overview. We begin with a query image and a domain-relevant prompt (left). The query is processed through textual inversion~\cite{gal2022image} with different token lengths. With each inversion, images are reconstructed and edited (generation with variations). After ensuring each reconstruction is in-distribution, we estimate the concept generative quality~\cite{fu2023dreamsim} (right).}
    \label{fig:method}
\end{figure*}

\paragraph{\textbf{Stable Diffusion}} The Stable Diffusion model generates images conditioned on textual input. Its architecture comprises the following key components: \textit{Variational Autoencoder (VAE)}, encodes input images into a latent space and decodes latent representations back into images. Let $x$ be an input image and $z$ be its latent representation. The encoding and decoding processes can be represented as:
\begin{equation}
z = \text{VAE}_{\text{Encoder}}(x), \quad x' = \text{VAE}_{\text{Decoder}}(z),
\end{equation}
where $x'$ is the reconstructed image.
\textit{Text Encoder} converts input text $t$ into embeddings ${e}_t$ that capture the semantic meaning of the text. These embeddings are used to condition the image generation process:${e}_t = \text{TextEncoder}(t).$
And a \textit{U-Net}, a convolutional neural network that operates on the latent space to modify the encoded image representation based on the text embeddings. It plays a crucial role in ensuring that the generated image aligns with the textual description.

The image generation process in Stable Diffusion involves encoding the input image into a latent representation using the VAE, conditioning this representation on the text embedding from the Text Encoder, and then refining this conditioned representation using the U-Net. Finally, the refined representation is decoded back into an image using the VAE decoder, resulting in an output image that is both realistic and semantically aligned with the input text.

\paragraph{\textbf{Textual Inversion}}
Textual inversion is a method employed in T2I latent diffusion models, such as Stable Diffusion, to adapt the model for generating images that are specific to a particular visual concept or object, which may not have been present in the original training data. This is achieved by fine-tuning a pre-trained T2I model on a selected set of images ${x_1, x_2, \ldots, x_n}$ that represents the target concept. The fine-tuning process results in the creation of a distinctive token $S^{*}$, which encapsulates the visual characteristics of the concept. During the training process, only the parameters associated with the text embeddings are updated, while the parameters of the VAE and the U-Net components of the model remain unchanged. This selective training approach ensures that the model retains its general image generation capabilities while learning to associate the new token $S^{*}$ with the specific visual attributes of the concept. Once trained, the token $S^{*}$ can be used in the text input of the T2I model to generate new images that exhibit the learned concept, allowing for controlled and targeted image synthesis. 

\subsection{Method}
Our approach builds on the textual inversion technique introduced by \cite{gal2022image}, which was originally designed for personalization and editing tasks by representing concepts with a single token. Unlike the original purpose of the method, our research aims to enhance the interpretability of the manifold of text-to-image (T2I) models, focusing on the originality of images rather than objects. For this purpose, we extend the method to employ \emph{multiple tokens}, building on the fact that a single-token representation may not sufficiently capture a complex, original, image. Overall, through our experiments, we find and demonstrate that the number of tokens required for reconstruction is correlated with the originality of an image. Thus, we use the number of tokens as a measure of originality.

\paragraph{Single Token vs. Multi-Token:} Let $T$ be a set of tokens representing a concept, where $T = {t_1, t_2, \dots, t_m}$ is a multi-token representation with $m$ tokens. In the original textual inversion method, a single token ($m=1$) is used to represent a concept. In contrast, our extension allows for multiple tokens ($m>1$) to represent the concept in a more detailed manner. The representation of the concept in the latent space can be expressed as a sequence of the embeddings of the tokens: \sm$= e_t(t_1) \space {e}_t(t_2), \dots, e_t(t_m)$, where $e_t$ is the embedding function and \sm, is the concatenated embedding of the tokens representing the query image. This multi-token approach enables a more granular exploration of the T2I model's manifold, facilitating a deeper understanding of the relationships between text and image representations, especially in the context of interpreting the originality of individual images. Following the creation of \sm, we can then use the original model to query with this new sequence \sm using the existing vocabulary of the text encoder.

Once we have the \sm sequence that represents the query image, the overall process involves two main steps: reconstruction and in-distribution evaluation.

\paragraph{\textbf{Reconstruction}}
To assess the quality of reconstruction, we employ the DreamSim score \cite{fu2023dreamsim}, which is a SOTA distance metric to measure the similarity between the generated image and the query image. For a given image represented by a set of tokens $T = {t_1, t_2, \dots, t_n}$, we generate a set of images ${x'_1, x'_2, \dots, x'_{20}}$, where each image $x'_i$ is generated using the textual inversion method with tokens $T$. The reconstruction score for each image is calculated as:
$\text{Reconstruction Score}(x'_i) = \text{DreamSim}(x'_i, x),$
where $x$ is the original query image. The overall reconstruction score for the concept is the average of the scores for the 20 generated images:
$\text{Average Reconstruction Score}(T) = \frac{1}{20} \sum_{i=1}^{20} \text{Reconstruction Score}(x'_i).$
Lower scores indicate better reconstruction and the results are plotted for visualization to provide a comprehensive understanding of the model's performance. Additional details on the DreamSim metric are provided in~\cref{sec:apx_dreamsim}.

\begin{table*}[ht]
\centering
\scalebox{0.85}{
{\normalsize  
\setlength{\tabcolsep}{3pt}
\begin{tabular}{>{\centering\arraybackslash}m{2cm} >{\centering\arraybackslash}m{2cm}  ccc|ccc}

\toprule
\textbf{Domain} & \textbf{Initial Tokens} & \multicolumn{3}{c}{\textbf{Common}} & \multicolumn{3}{c}{\textbf{Original}} \\ 
\midrule
Houses & \texttt{"house"}  & \adjustbox{valign=c}{\includegraphics[width=1.8cm, height=1.6cm]{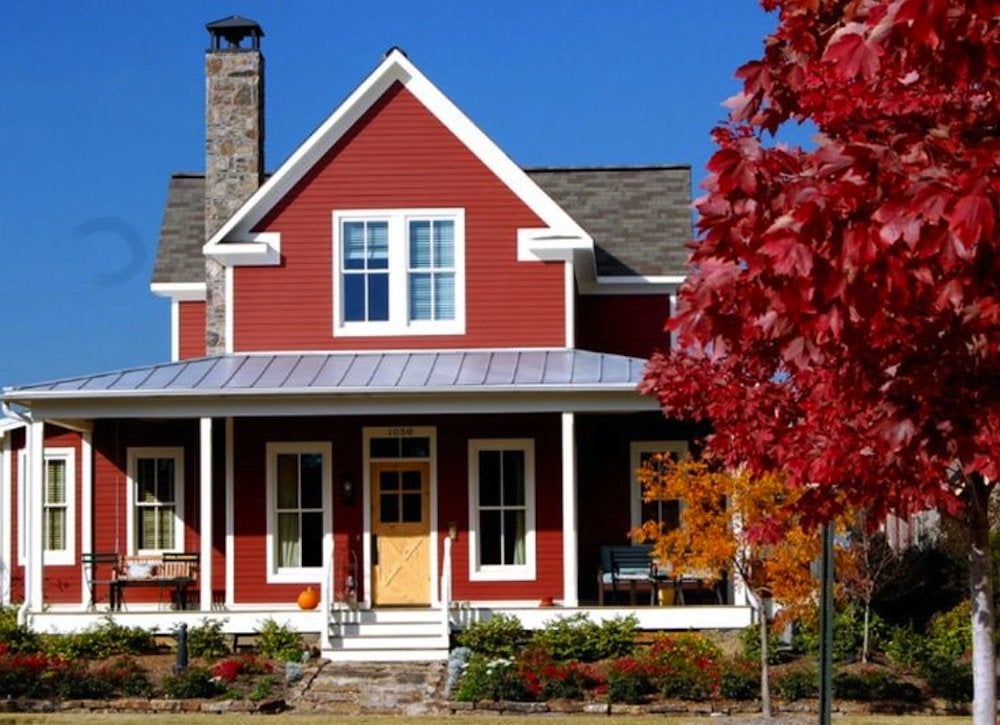}} & \adjustbox{valign=c}{\includegraphics[width=1.8cm, height=1.6cm]{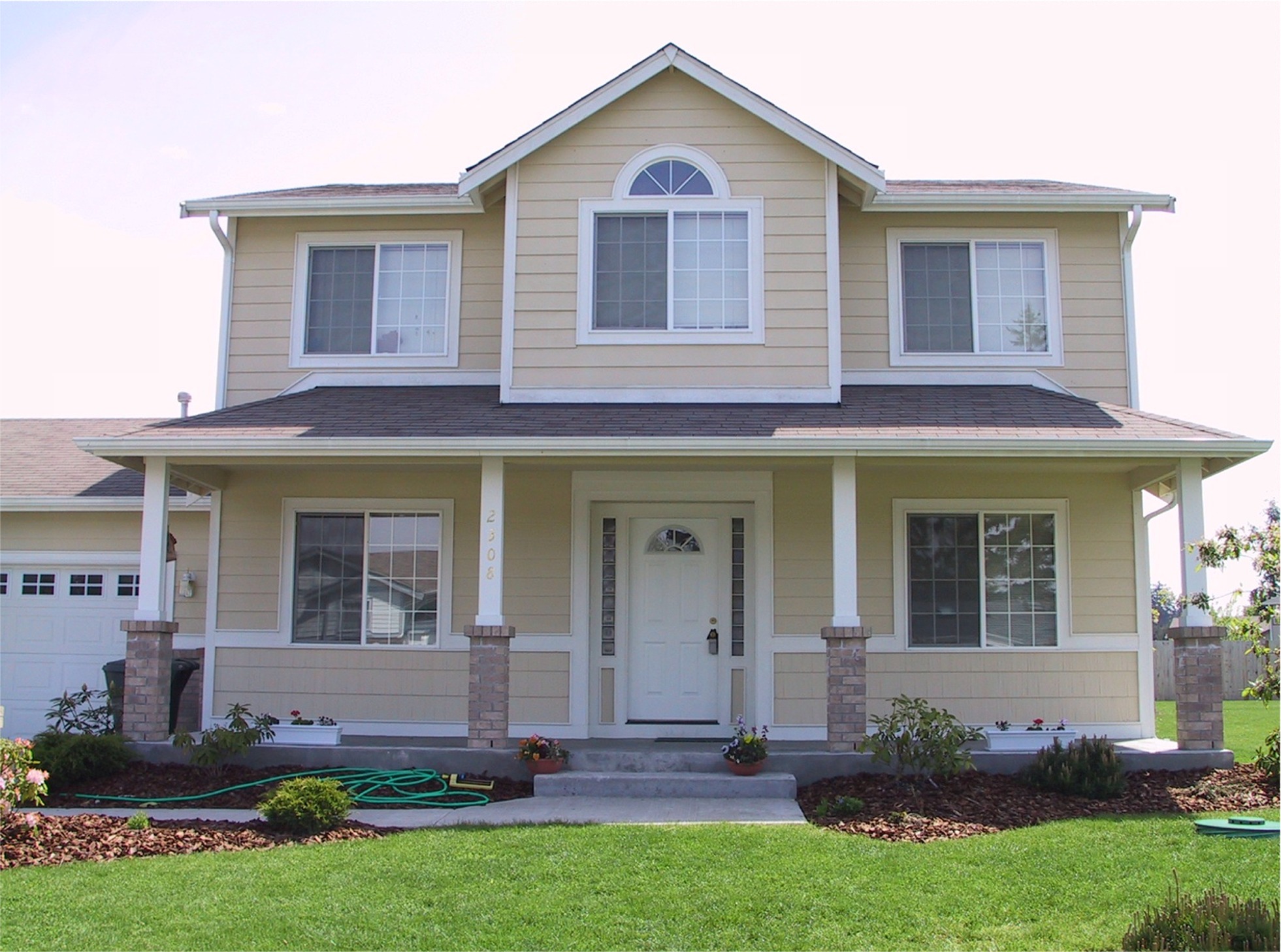}} & \adjustbox{valign=c}{\includegraphics[width=1.8cm, height=1.6cm, ]{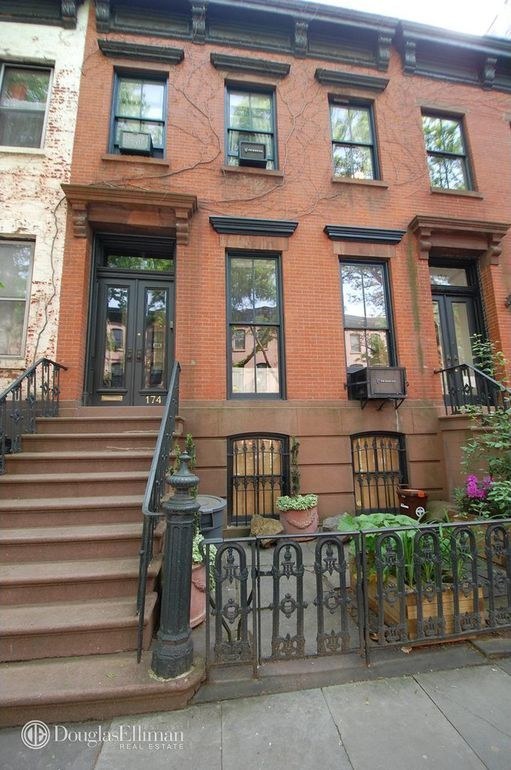}} \hspace{5pt} & \hspace{5pt} \adjustbox{valign=c}{\includegraphics[width=1.8cm, height=1.6cm, ]{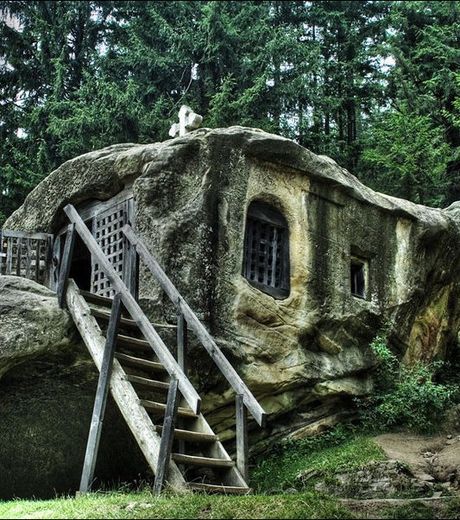}} & \adjustbox{valign=c}{\includegraphics[width=1.8cm, height=1.6cm, ]{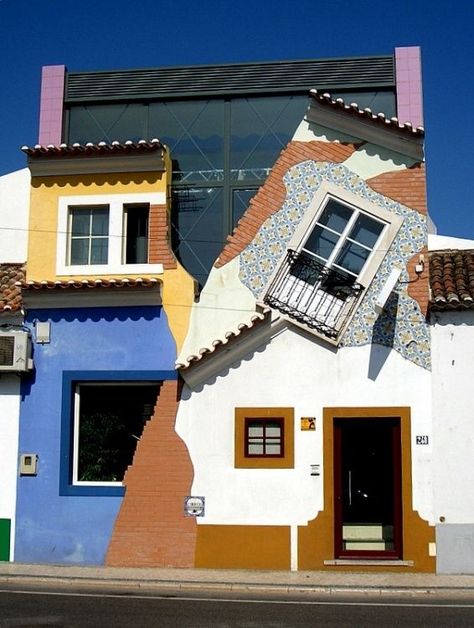}} & \adjustbox{valign=c}{\includegraphics[width=1.8cm, height=1.6cm, ]{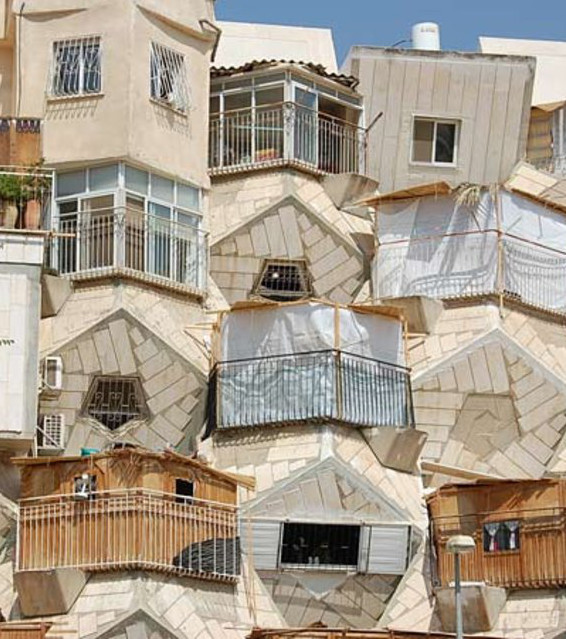}} \\ 
\addlinespace[0.3em]
Artwork & \texttt{"painting"}  &\adjustbox{valign=c}{\includegraphics[width=1.8cm, height=1.6cm, ]{figures/experimental_setup/real_data_samples/artwork_common1.jpg}} &\adjustbox{valign=c}{\includegraphics[width=1.8cm, height=1.6cm, ]{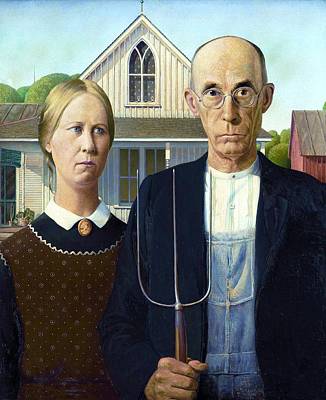}} &\adjustbox{valign=c}{\includegraphics[width=1.8cm, height=1.6cm, ]{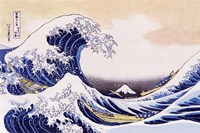}} \hspace{5pt} & \hspace{5pt} \adjustbox{valign=c}{\includegraphics[width=1.8cm, height=1.6cm, ]{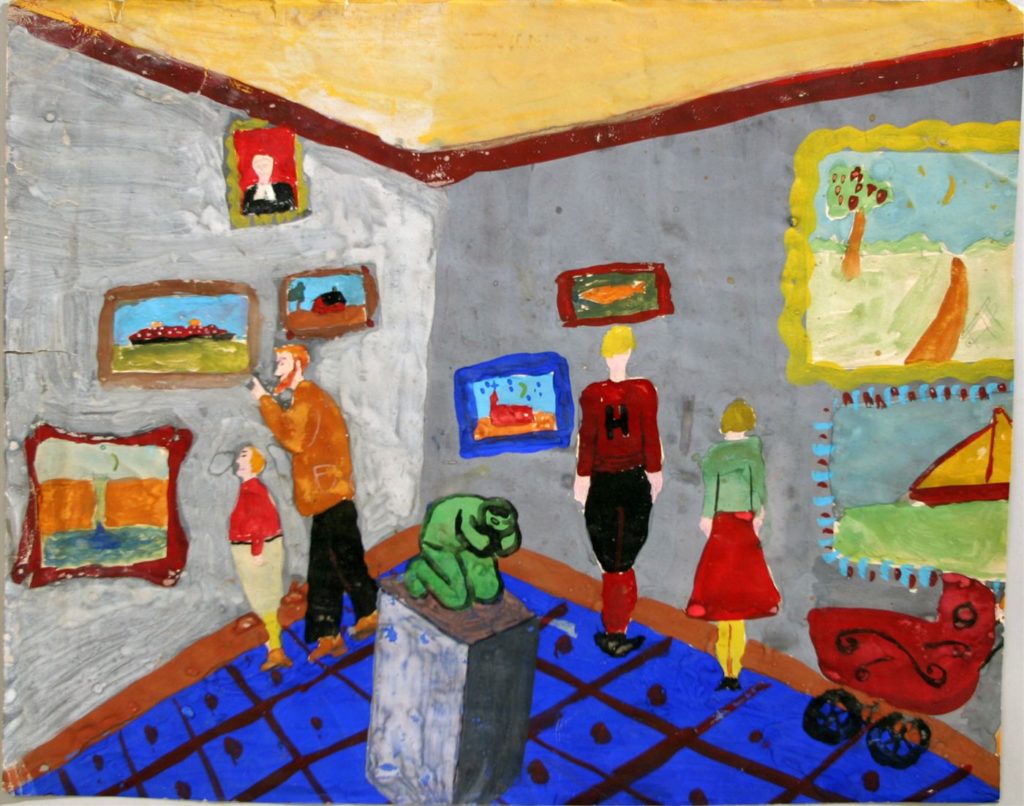}} &\adjustbox{valign=c}{\includegraphics[width=1.8cm, height=1.6cm, ]{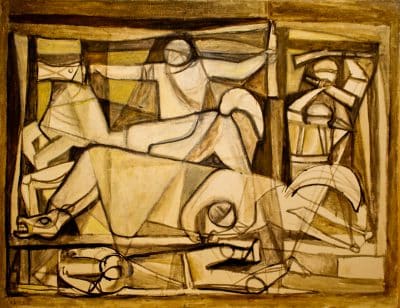}} &\adjustbox{valign=c}{\includegraphics[width=1.8cm, height=1.6cm, ]{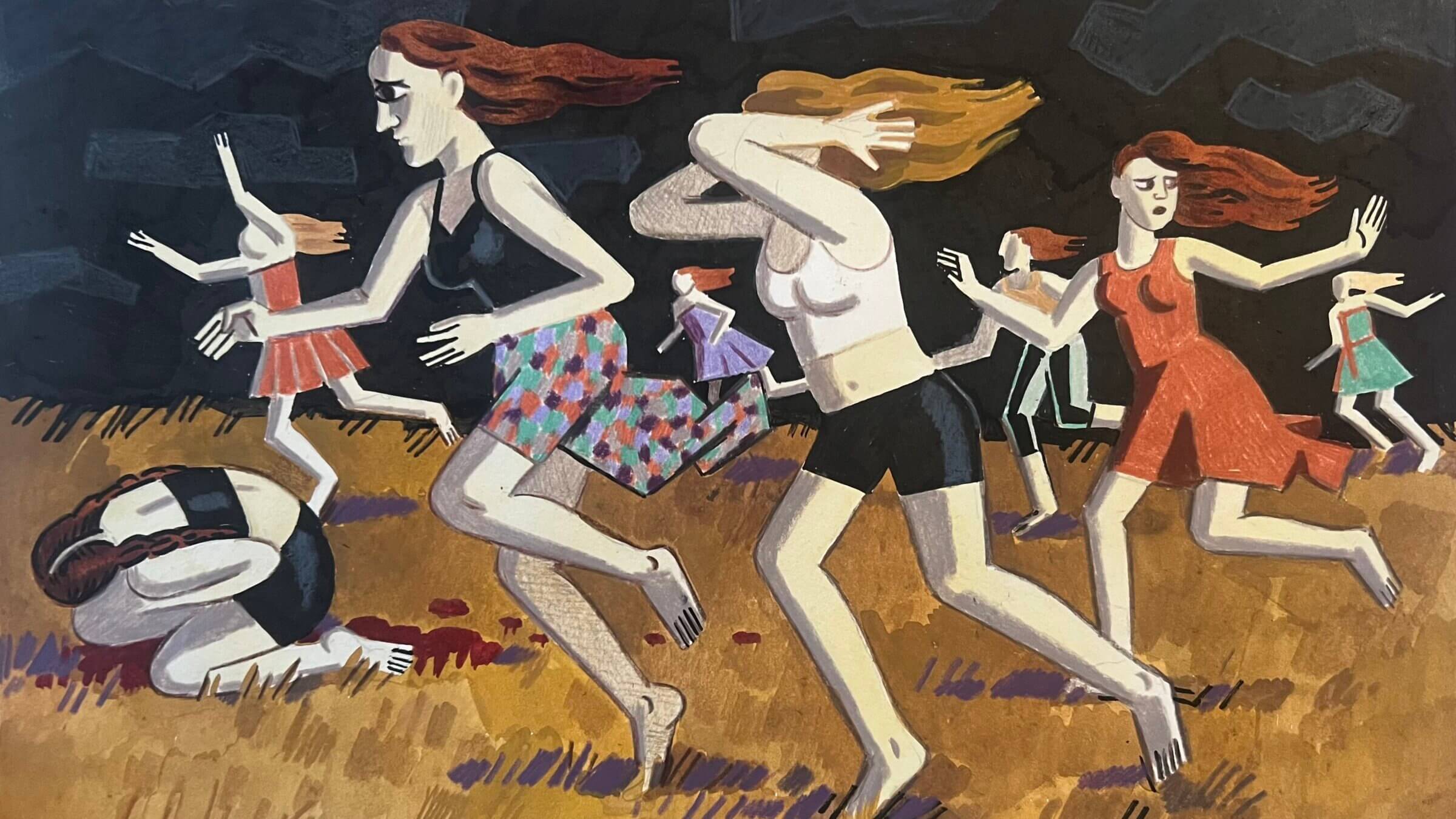}} \\
\addlinespace[0.3em]
Animals & \texttt{"dog" "cat" "bird"}  &\adjustbox{valign=c}{\includegraphics[width=1.8cm, height=1.6cm, ]{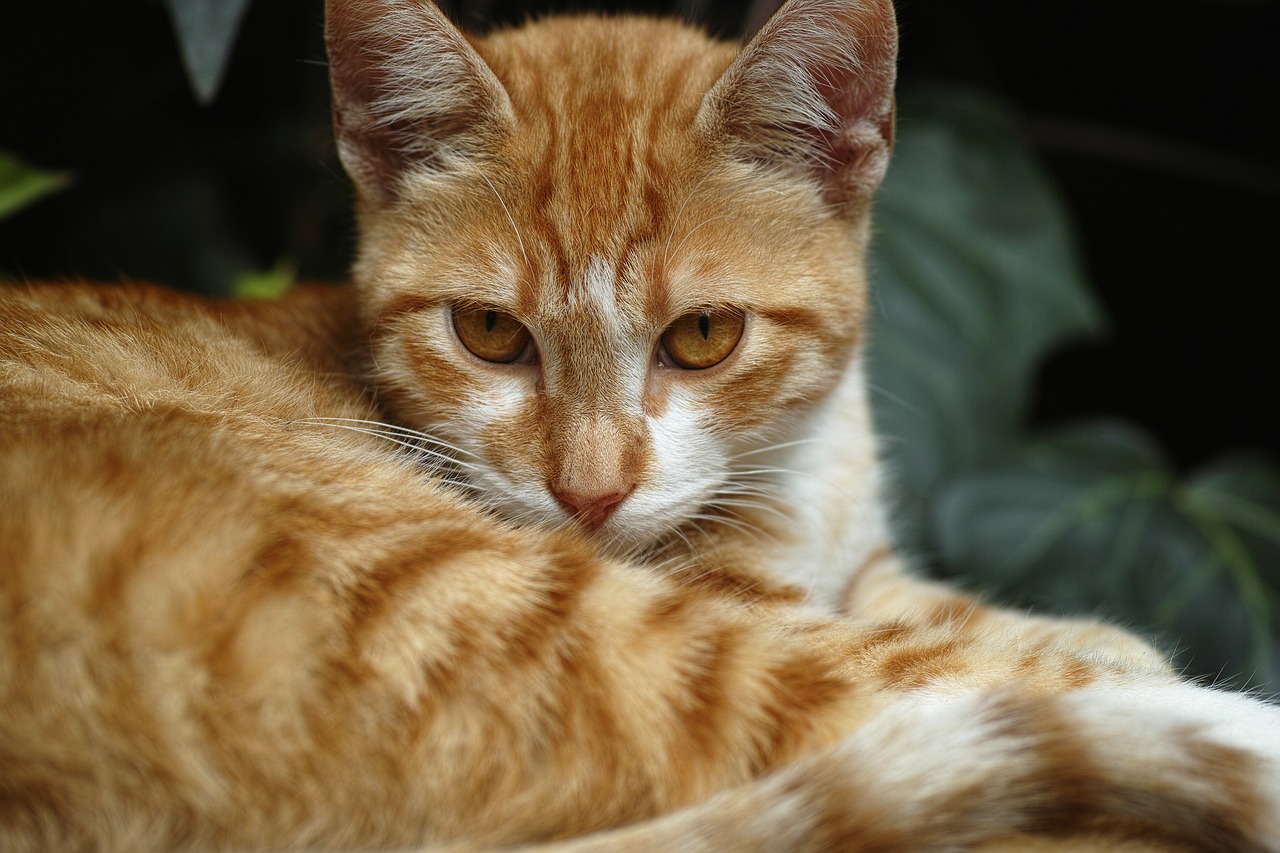}} &\adjustbox{valign=c}{\includegraphics[width=1.8cm, height=1.6cm, ]{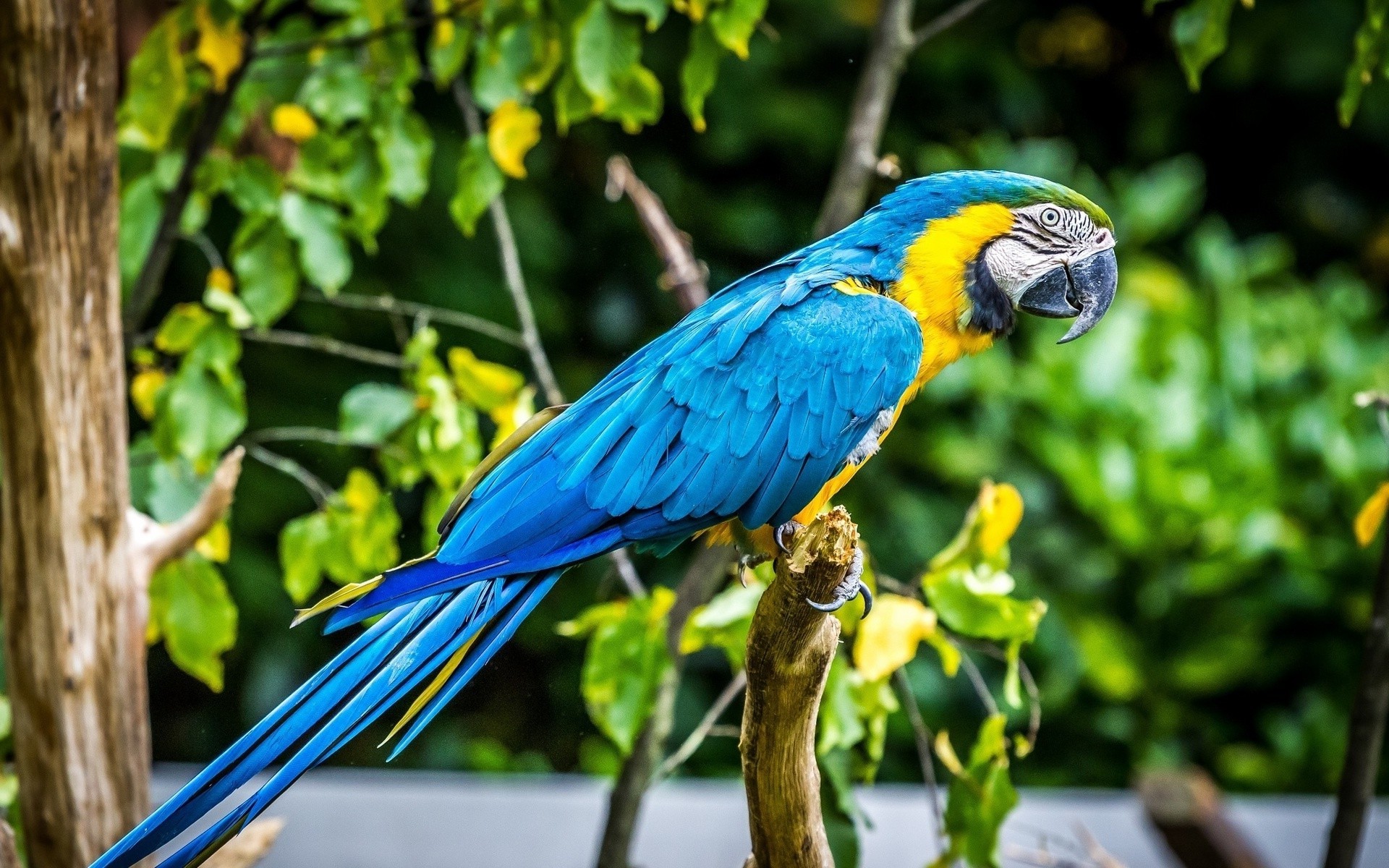}} &\adjustbox{valign=c}{\includegraphics[width=1.8cm, height=1.6cm, ]{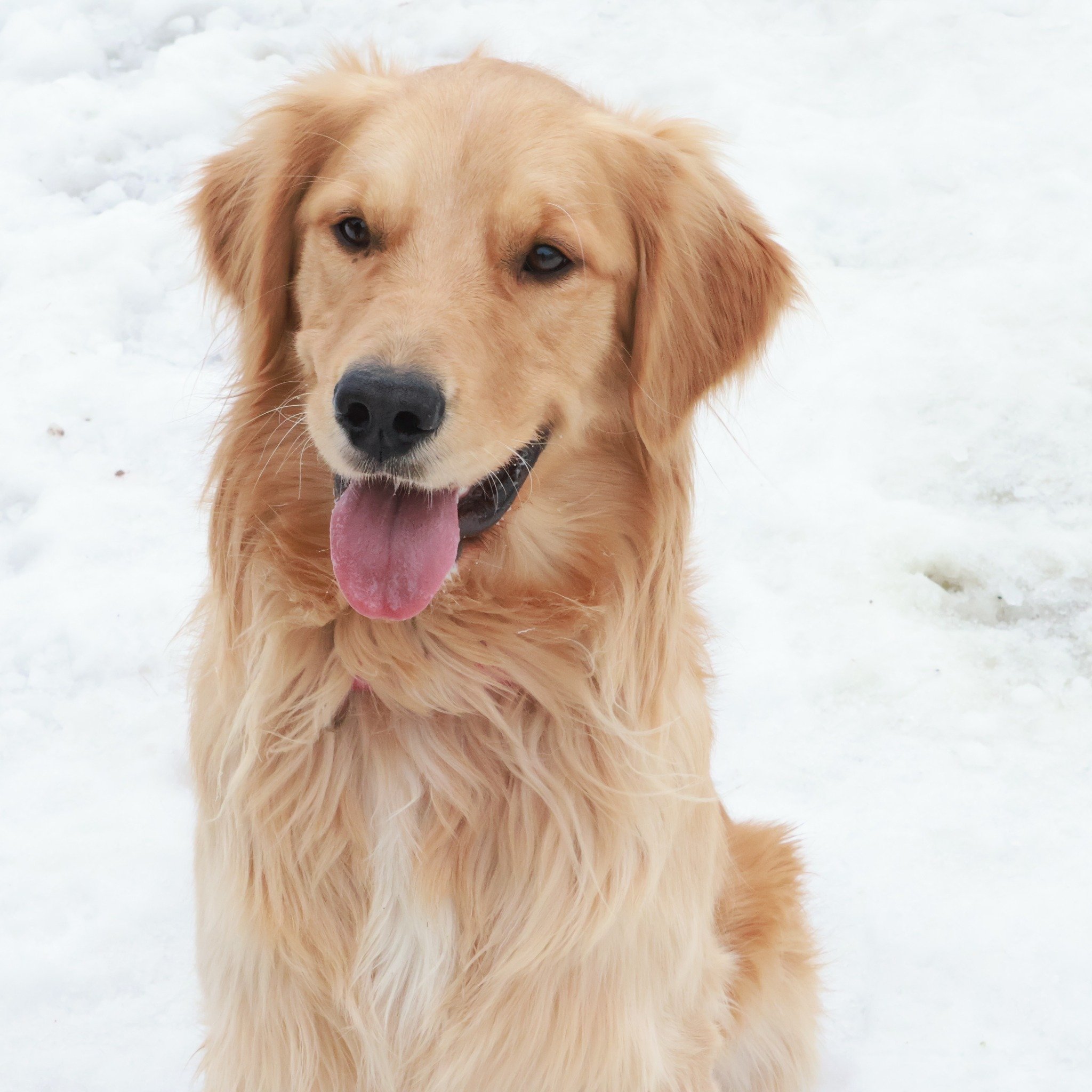}} \hspace{5pt} & \hspace{5pt} \adjustbox{valign=c}{\includegraphics[width=1.8cm, height=1.6cm, ]{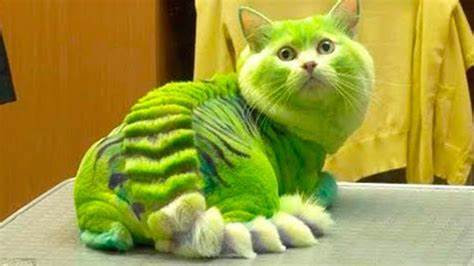}} &\adjustbox{valign=c}{\includegraphics[width=1.8cm, height=1.6cm, ]{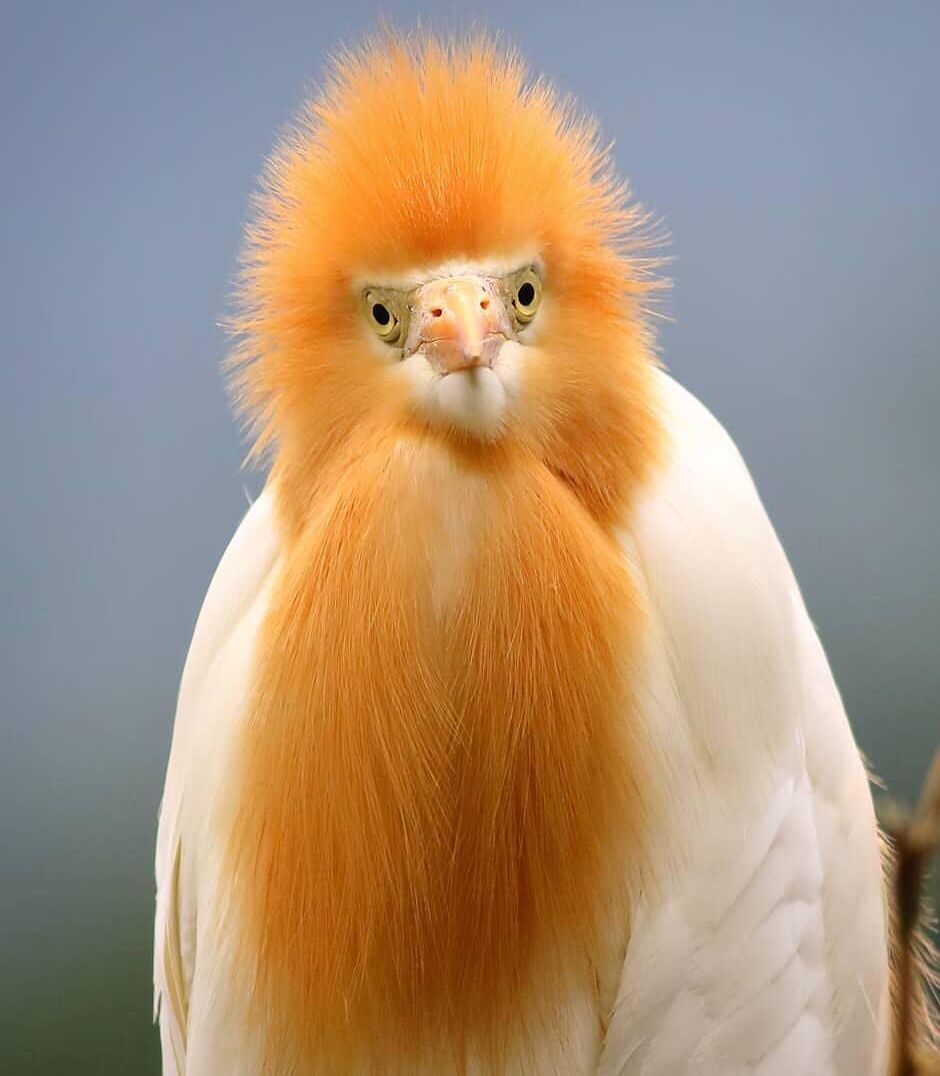}}&\adjustbox{valign=c}{\includegraphics[width=1.8cm, height=1.6cm, ]{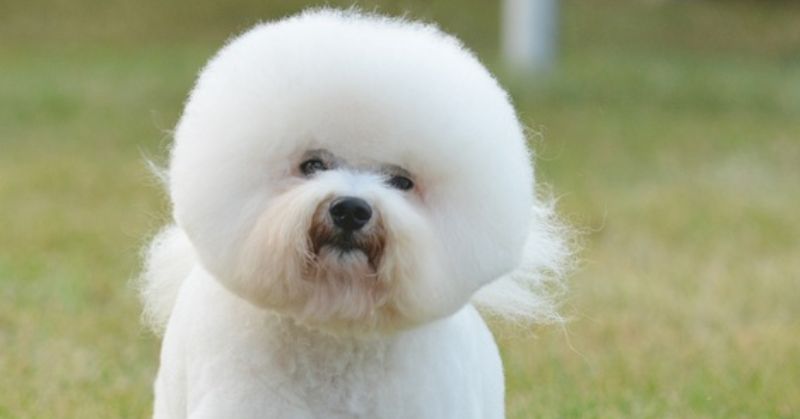}} \\
\addlinespace[0.3em]
Sport Photography & \texttt{"sport"}  &\adjustbox{valign=c}{\includegraphics[width=1.8cm, height=1.6cm, ]{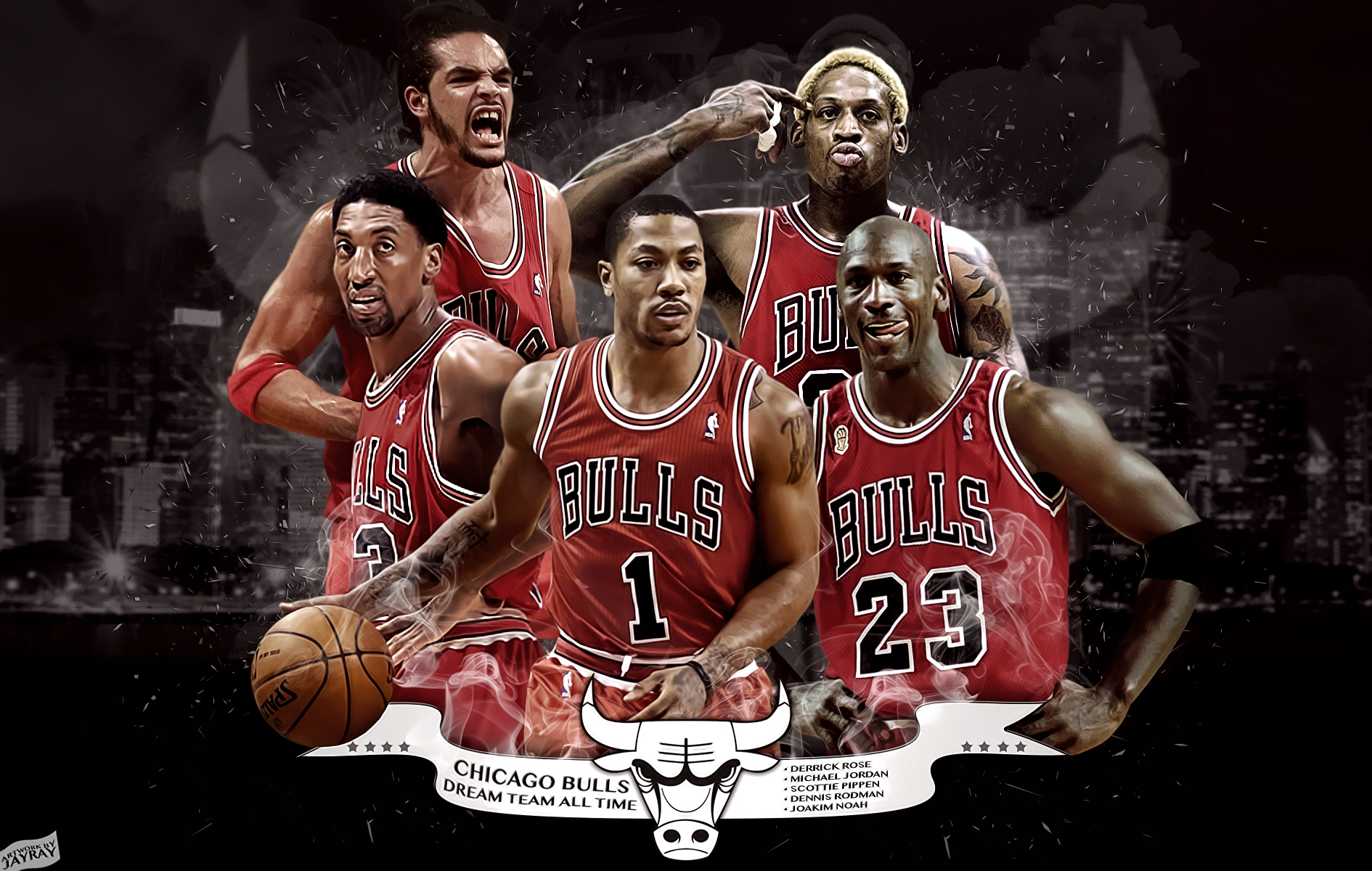}} &\adjustbox{valign=c}{\includegraphics[width=1.8cm, height=1.6cm, ]{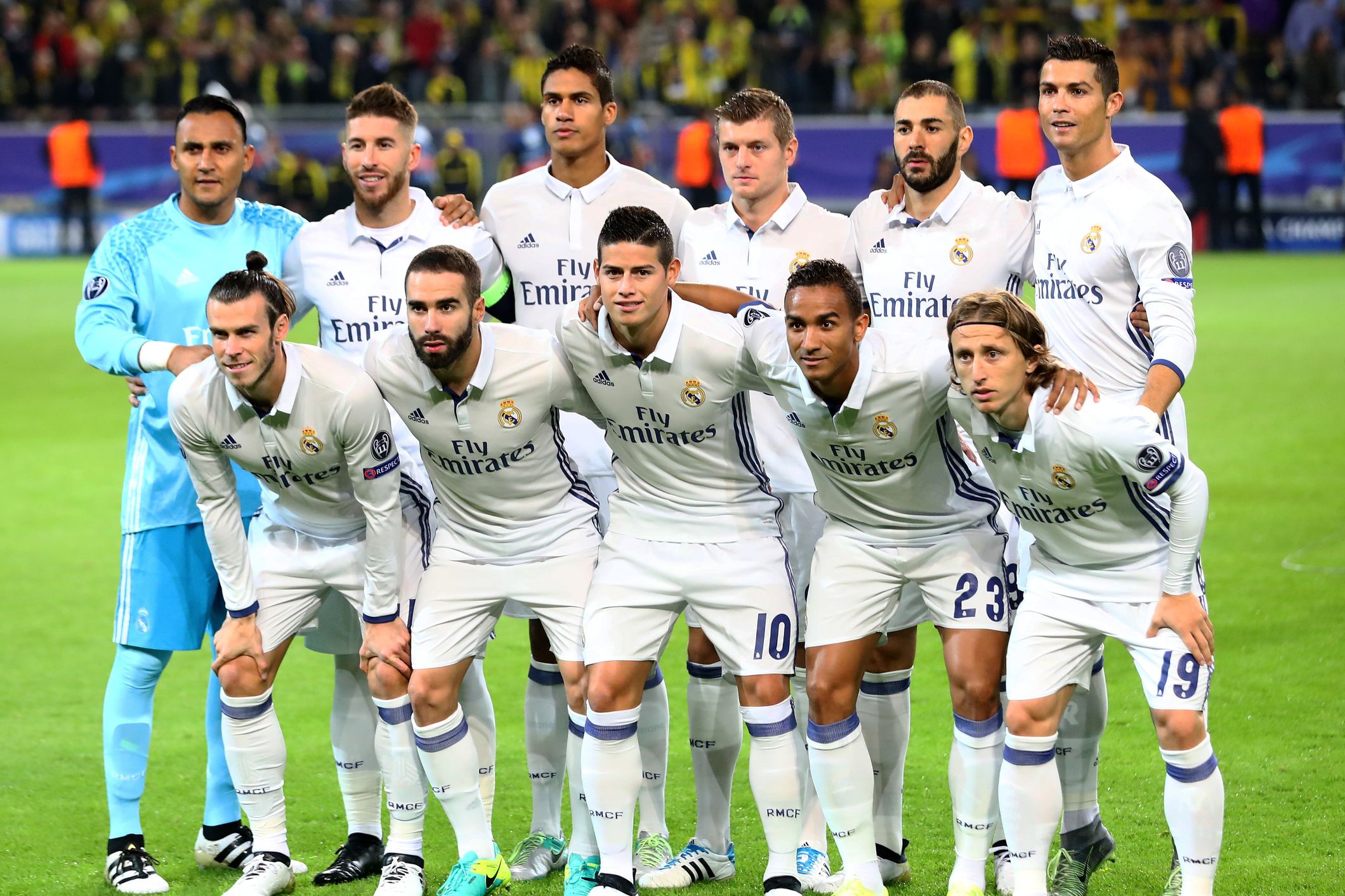}} &\adjustbox{valign=c}{\includegraphics[width=1.8cm, height=1.6cm, ]{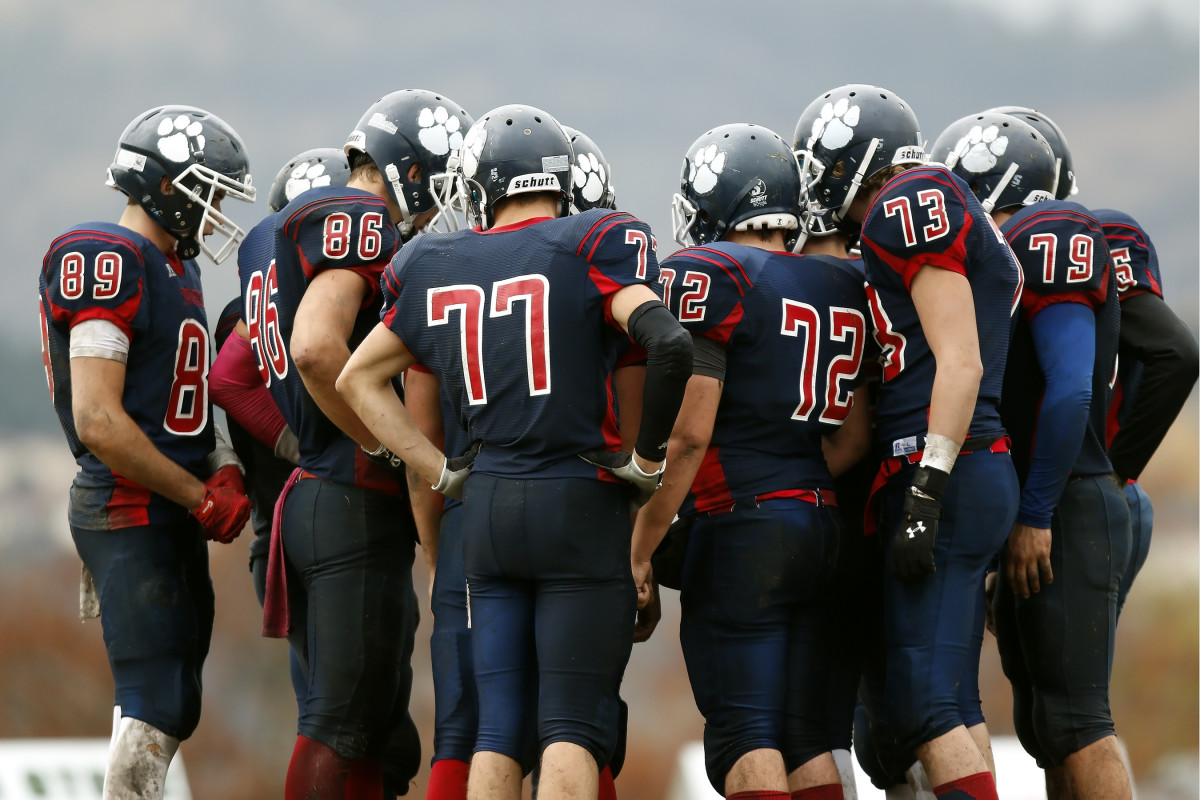}} \hspace{5pt} & \hspace{5pt} \adjustbox{valign=c}{\includegraphics[width=1.8cm, height=1.6cm, ]{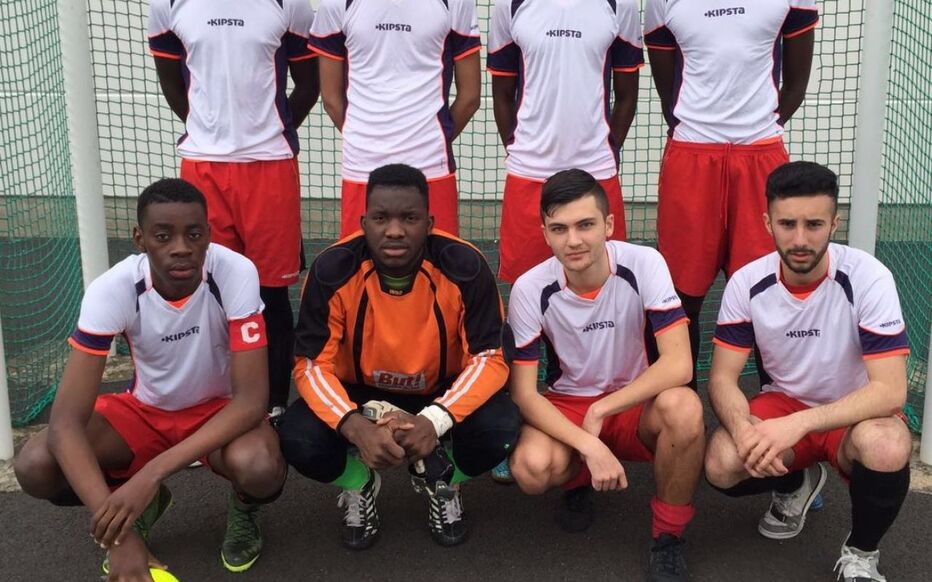}} & \adjustbox{valign=c}{\includegraphics[width=1.8cm, height=1.6cm, ]{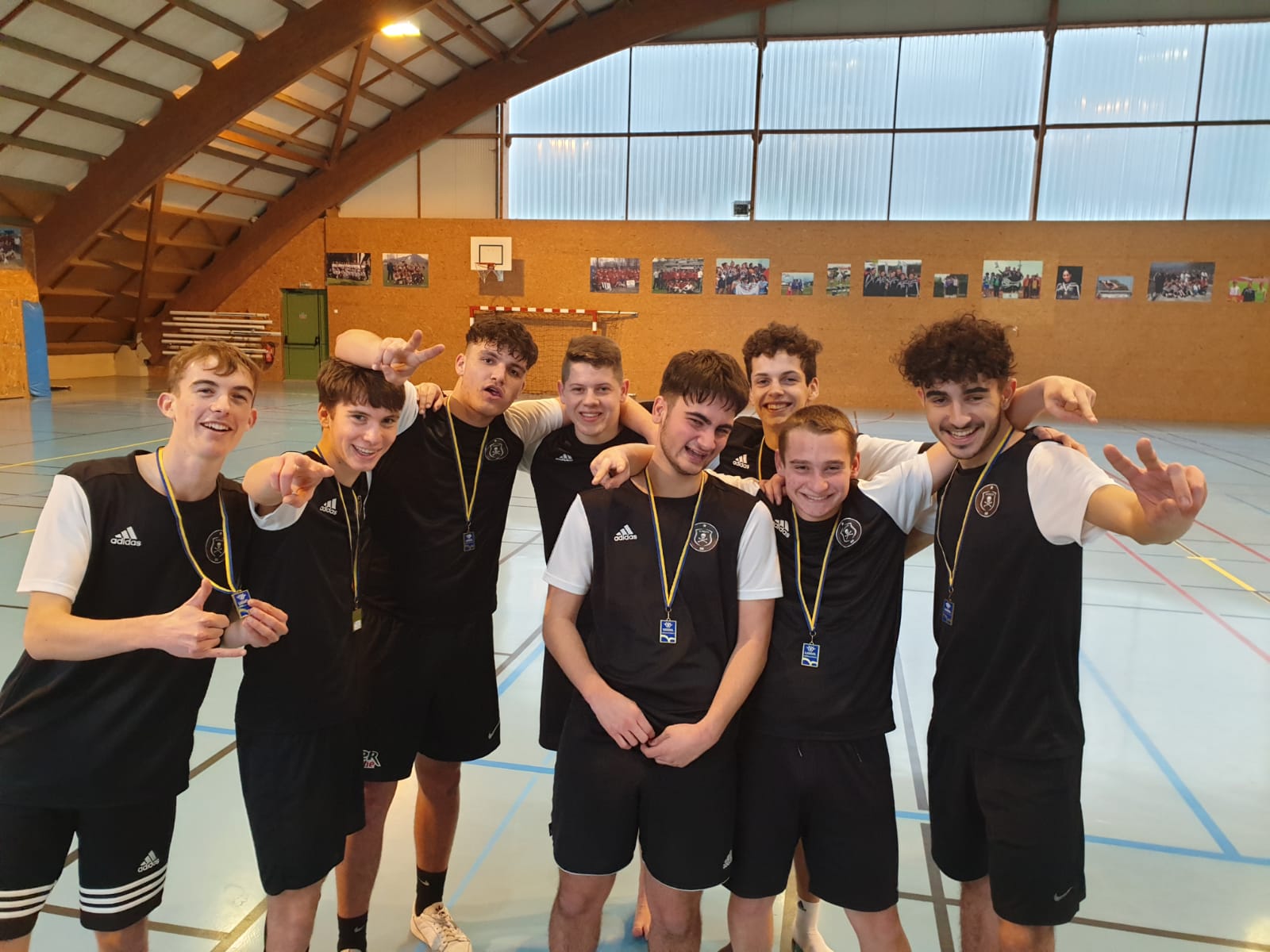}} &\adjustbox{valign=c}{\includegraphics[width=1.8cm, height=1.6cm, ]{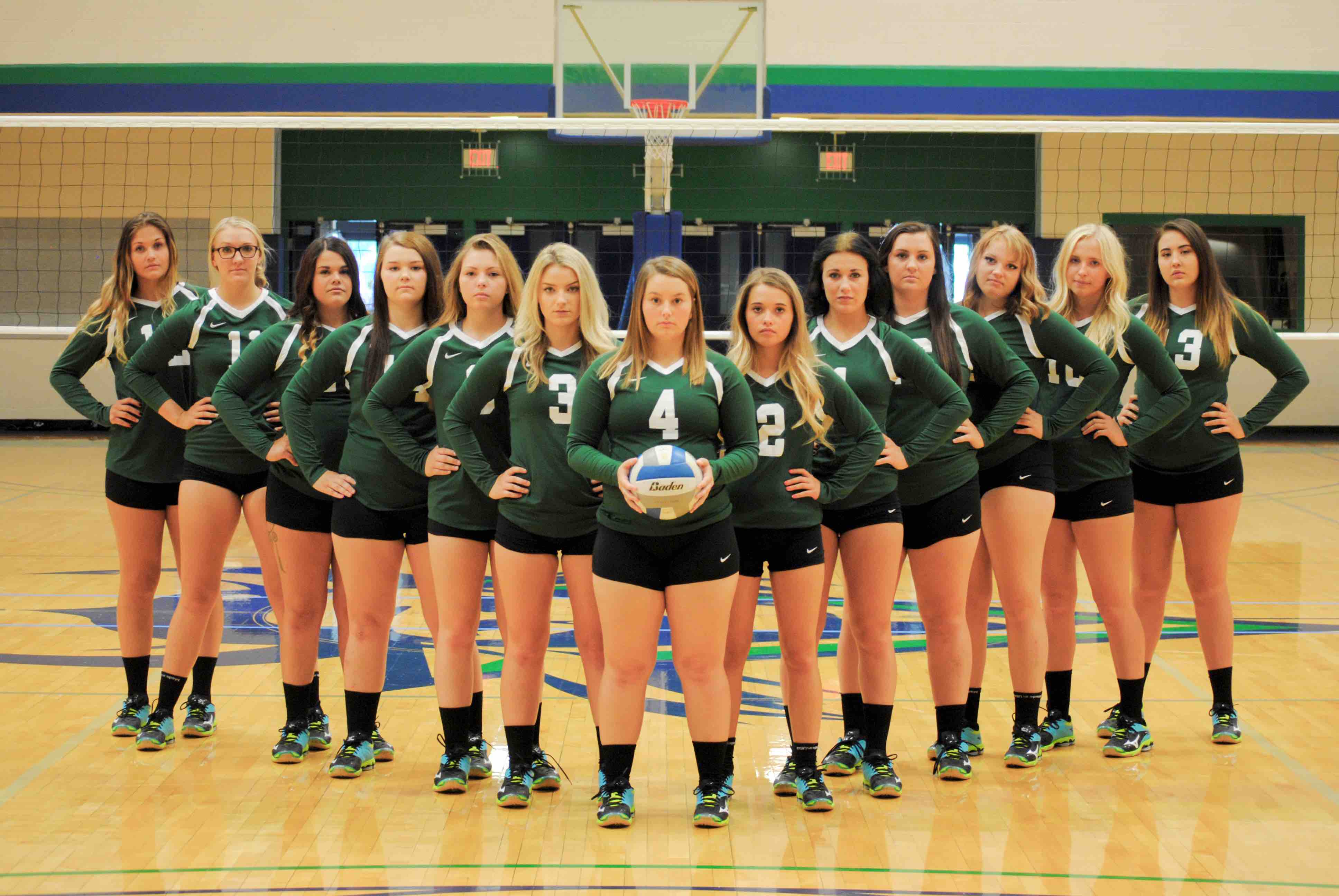}} \\
\addlinespace[0.3em]
People & \texttt{"person"}  &\adjustbox{valign=c}{\includegraphics[width=1.8cm, height=1.6cm, ]{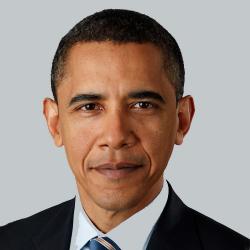}} &\adjustbox{valign=c}{\includegraphics[width=1.8cm, height=1.6cm, ]{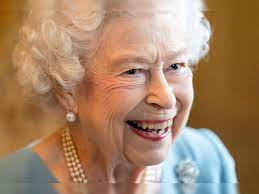}} &\adjustbox{valign=c}{\includegraphics[width=1.8cm, height=1.6cm, ]{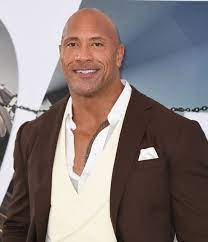}} \hspace{5pt} & \hspace{5pt} \adjustbox{valign=c}{\includegraphics[width=1.8cm, height=1.6cm, ]{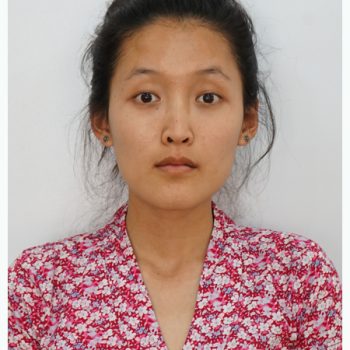}} &\adjustbox{valign=c}{\includegraphics[width=1.8cm, height=1.6cm, ]{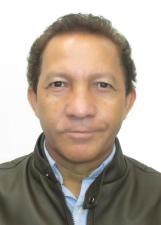}} &\adjustbox{valign=c}{\includegraphics[width=1.8cm, height=1.6cm, ]{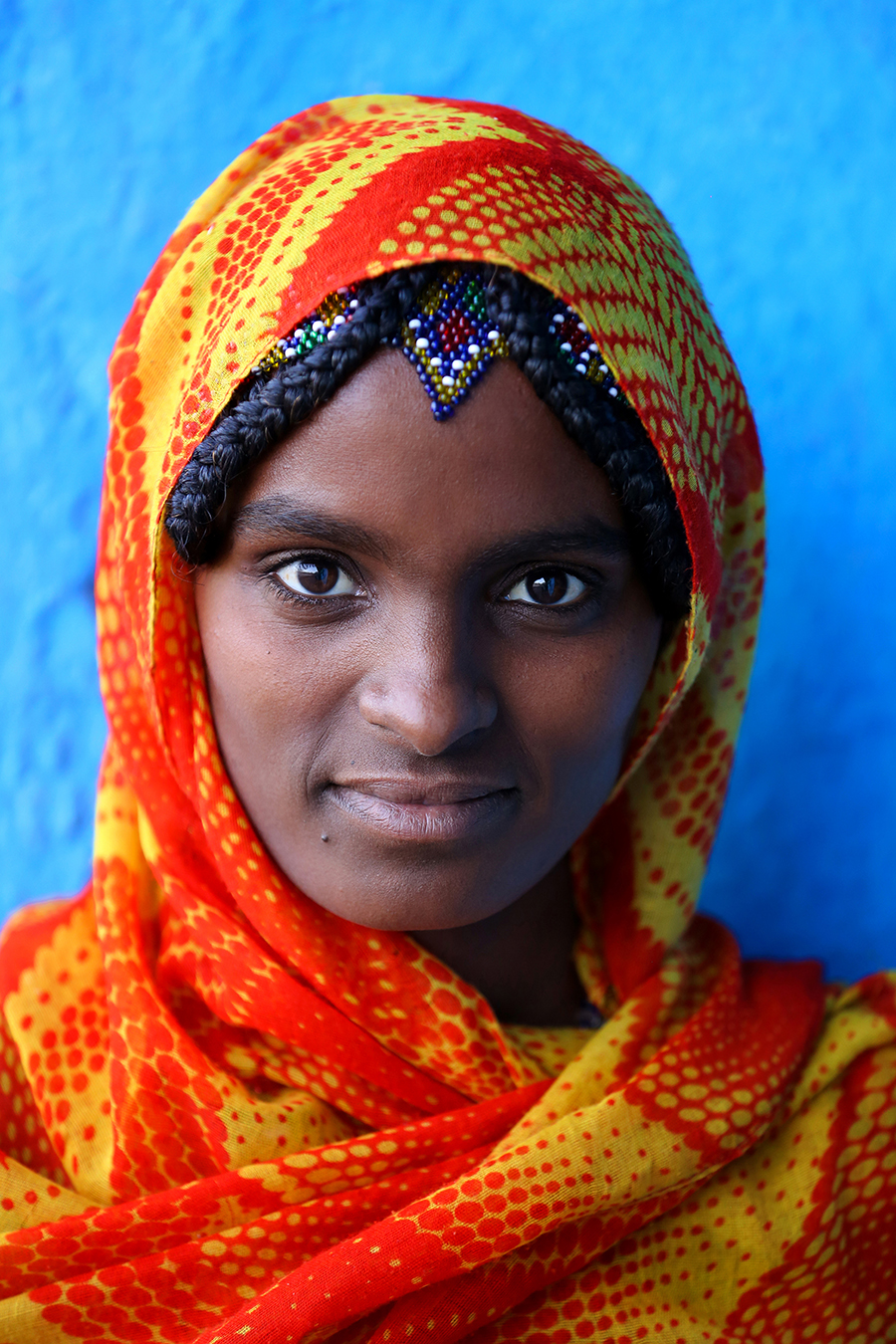}} \\
\bottomrule
\end{tabular}
} 
}
\vspace{0.5cm}
\caption{Sample Images from the dataset curated with samples of Common and Original images for each Domain}
\label{tab:ti_data_examples}

\end{table*}

\paragraph{In-Distribution Assessment}
We employ our method in two experimental setups, one synthetic playground for controlled occurrence distributions and their induced behavior and the other for the real-world scenario.
For \textit{Real-world} Settings, we use editability as the criterion to maintain in-domain generation \cite{gal2022image,gal2023designing}. Specifically, we use prompts like "cat in \sm" to generate images that are both representative of the concept in the query image and editable within the domain of the model. For the \textit{synthetic} settings, due to the model's simplicity of the data it was trained on, editability is not necessarily the right measure of in-domain generation. However, The underlying advantage of the synthetic data is that the distribution of the data is fully known. This allows us to measure in-domain generation by a more informed measure which we can validate. Instead of editability, we check that for every seed, the shapes are in different positions, as the data distribution positioned the shapes randomly (by design).  This validation ensures that the model has not reached a point of overfitting. We provide an ablation study in~\cref{sec:apx_syn_indistribution_ablation} to justify this decision.

Finally, we assess the originality of the query images by combining the reconstruction and in-distribution validation. An illustration of our method is provided in Fig.~\ref{fig:method}. The combination of multi-token textual inversion and these evaluation criteria enables a more detailed and original content generation, contributing to the assessment of the originality of imagery and Interpretability of T2I diffusion models.

\section{Experimental setup}
\label{sec:exp}

We conduct our experiments on two main environments. Both are based on the architecture of the stable diffusion model and differ in the data they were trained on. The first is the controlled environment, which we trained from scratch on a controlled synthetic data ,detailed in \cref{sec:analysis}, and the second is the known public pretrained stable diffusion \footnote{we based our implementation on the Huggingface Diffusers framework at \url{https://huggingface.co/docs/diffusers/en/index}, and used its pretrained stable diffusion models.}.

\subsection{Synthetic Framework}
Our first set of experiments is conducted on the synthetic dataset described in \cref{sec:analysis}. As discussed, within this synthetic framework, we are able to provide evidence for generalization and hence validate or assumptions underlying our method for quantifying originality.

To conduct the experiment, separate Stable Diffusion models were trained on synthetic datasets as depicted in \cref{sec:analysis} (further details are provided in \cref{sec:appx_syn_framework}). We used a pretrained VAE and UNet from scratch and employing BERT as the Text-Encoder. For evaluation, a YOLOv8 model is fine-tuned on the synthetic datasets to detect and classify elements in generated images, ensuring high-quality detections with a confidence threshold of 0.9. Additional details are described in~\Cref{sec:appx_syn_framework}.

\paragraph{\textbf{Quantifying Originality in Synthetic Framework}}
In the context of assessing the originality of query images, we synthesize a custom dataset characterized by a non-trivial distribution. This dataset features three distinct element combinations, varying in occurrence frequency within the dataset, differing by orders of magnitude. Notably, as the position of each element in the image is randomly assigned, a higher occurrence frequency within the dataset signifies greater genericity. We validate our methodology within this controlled setting.

\subsection{Real-World Setting}
In the more elaborate setting, we used the widespread public Stable diffusion model\footnote{The pretrain Stable diffusion model was taken from \url{https://huggingface.co/CompVis/stable-diffusion-v1-4}}. We demonstrate our method on diverse domains, including houses, artwork, sports, animals and people's faces. For each, we initialize all tokens in the learned prompt with a relevant initial token, train the model to discover tokens for the query image, and measure editability by validating the existence of a cat when prompted with "cat in \sm." List of the domains evaluated and examples of the curated data is provided in ~\cref{tab:ti_data_examples}.

\paragraph{\textbf{Implementation Details}}
We trained a multi-token textual inversion variant with different sizes of token length, starting from 1 and up to 5 consecutive tokens in the sequence. The training was conducted with a batch size of 20, a learning rate of $5e-4$, and a total of 2000 steps, using 35 denoising inference steps. Further details, including the training prompts and the training scheme, are provided in~\cref{sec:apx_text_inversion_impl}.
\section{Results}
\label{sec:results}
In this section, we present the results from our experiments, described in the previous section, demonstrating that original images require more tokens for more accurate representation. Additional Results, including demonstrations of the In-Distribution assessment, are provided in~\Cref{sec:app_additional_results}.

\paragraph{Synthetic experiments}

\begin{figure}[H]
    
    \centering
\includegraphics[width=0.95\linewidth]{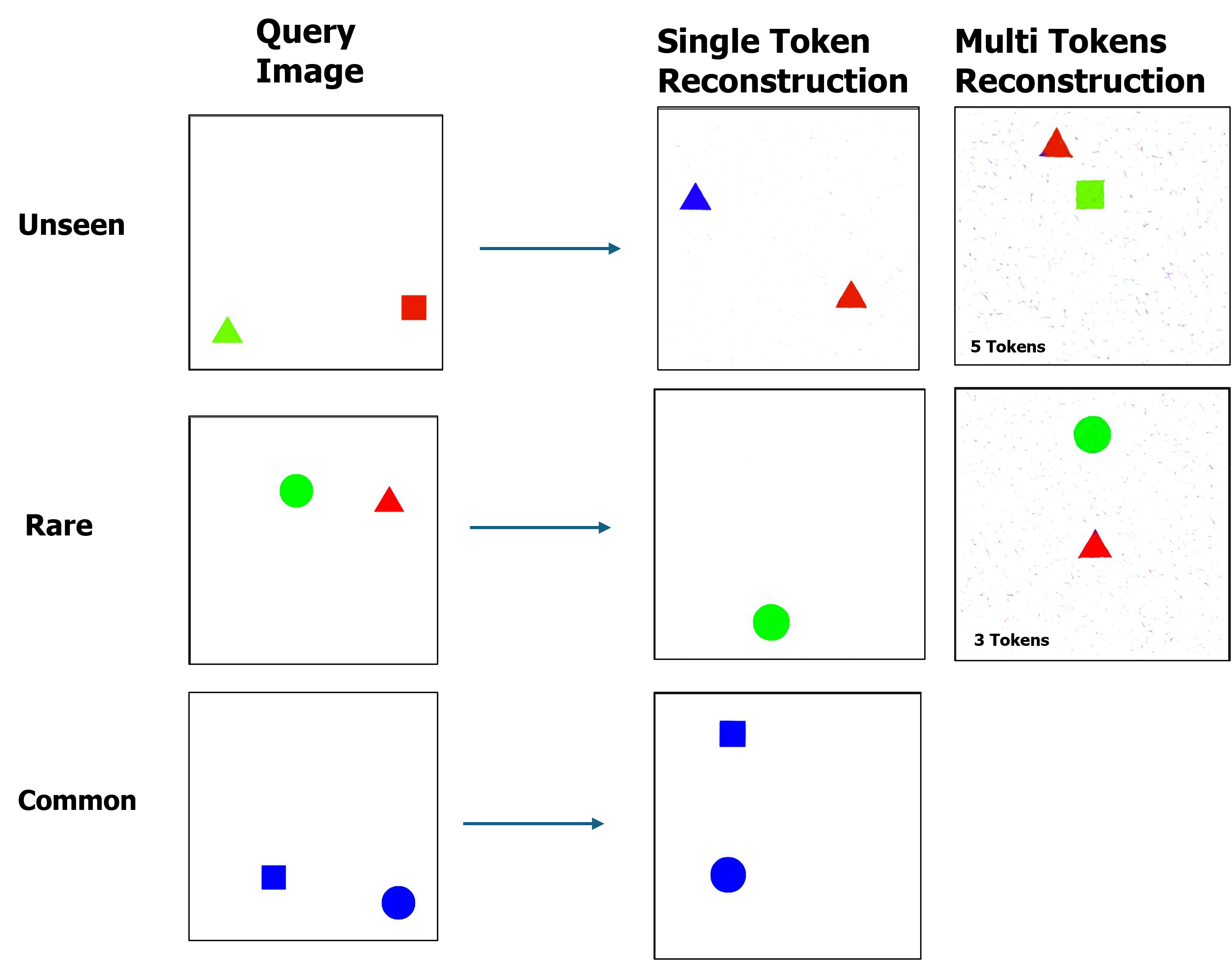}
    \caption{Qualitative results for reconstructing common, rare, and unseen images. Unseen concepts require five tokens for correct reconstruction, rare images require three, and common ones only one.} 
    \label{fig:res_syn}
    \vspace{-0.3cm}
\end{figure}

We run our evaluation method also in the synthetic setting, as qualitatively depicted in Fig.~\ref{fig:res_syn}. Supporting the real-world experiments, here, too, we observe a spectrum of behaviors, spanning from familiar concepts requiring only a single token (bottom), through rare examples that require three tokens (middle), to unseen concepts that require five for correct reconstruction (top).
\begin{figure}[t]
    \centering
    \includegraphics[width=.99\linewidth]{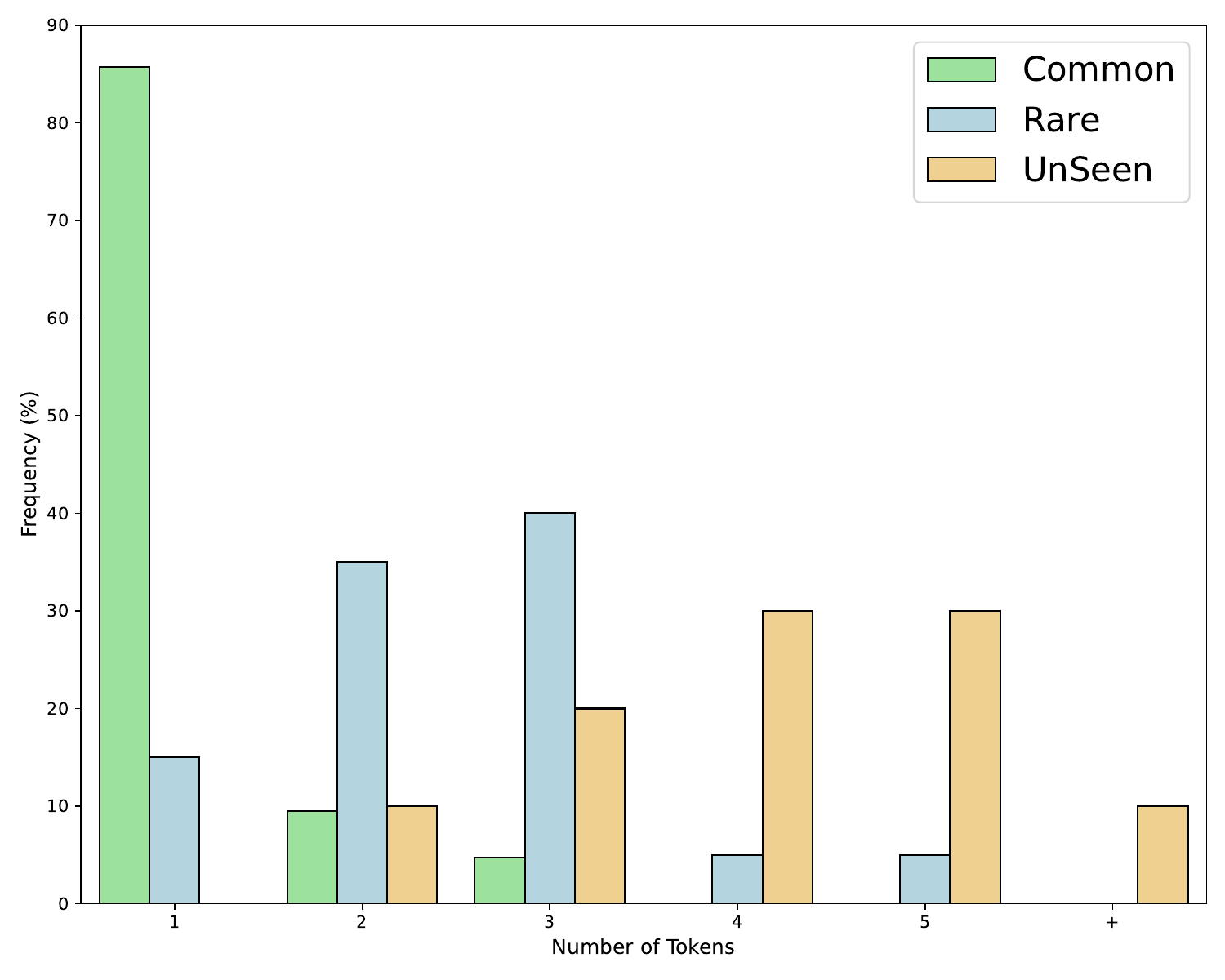} 
     \vspace{-0.1cm}
    \caption{Minimum number of tokens required for image Reconstruction of Data by frequency in the training data in the Synthetic setting. The plot illustrates the distribution of the minimum number of tokens needed to reconstruct original images from three categories: Common, Rare, and Unseen.}
    \vspace{-0.1cm}
    \label{fig:sup_syn_ti}
\end{figure}

We also conduct a quantitative analysis, the results of which are depicted in Fig.~\ref{fig:sup_syn_ti}. In this experiment, we randomly selected 20 images from each of the three groups (Common, Rare, and Unseen). We then plotted the minimum number of tokens required to reconstruct the original images for each sample. The results indicate that for the majority of images categorized as Common, a mere single token was sufficient for reconstruction. In contrast, images classified as Rare typically necessitated between 2 to 3 tokens, while those labeled as Unseen generally required between 4 to 5 tokens for successful reconstruction. Images that could not be reconstructed within the 5-token constraint were denoted as "+" on the x-axis.

 This experiment underscores the varying degrees of sequence length required for reconstructing images across different categories. It provides a quantitative foundation for our approach to measuring the originality of a concept by quantifying its familiarity with the model.

\paragraph{Pretrained Stable Diffusion}
\begin{figure*}[h]
    \centering
\includegraphics[width=0.9\linewidth]{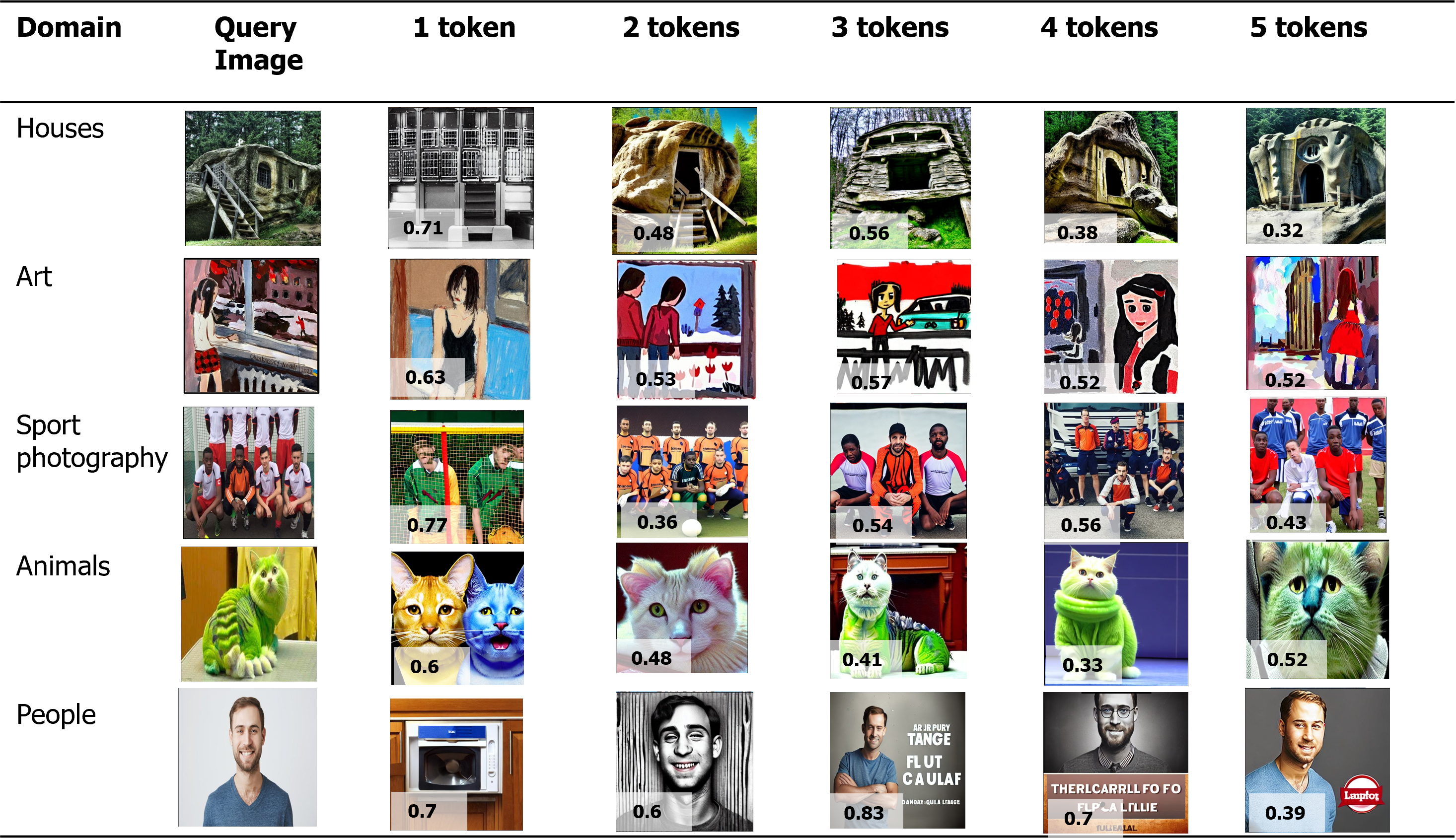}
    \vspace{-0.2cm}
    \caption{Qualitative results for reconstructing original images from various domains using multi-token textual inversion, demonstrating that for original images, more tokens improve capturing additional details of the query image. The average DreamSim score for each experiment is depicted at the bottom of each representative image.}
    \label{fig:res_stable_original}
    \vspace{-0.3cm}
\end{figure*}
\begin{figure*}[h]
    \centering
\includegraphics[width=0.95\linewidth]{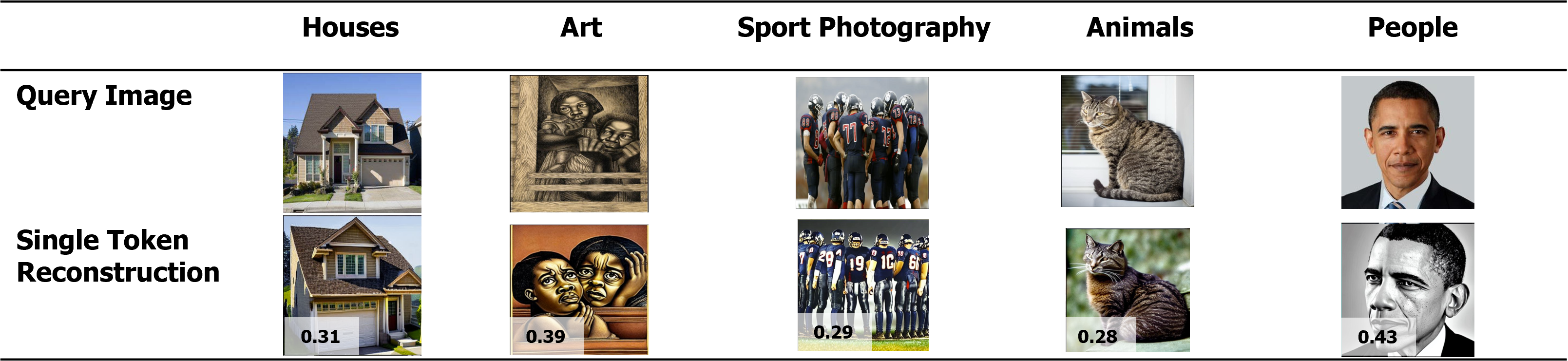}
    \vspace{-0.2cm}
    \caption{Qualitative results for reconstructing common images from various domains using multi-token textual inversion, demonstrating that for common images, a single token can reach high reconstruction capabilities. The average DreamSim score for each experiment is depicted at the bottom of each representative image.}
    \label{fig:res_stable_common}
    \vspace{-0.3cm}
\end{figure*}
Qualitative examples for reconstruction using textual inversion for original images (as labeled by a human expert) are provided in~\cref{fig:res_stable_original} and for common images in~\cref{fig:res_stable_common}. As seen, semantic preservation improves with the addition of more tokens for original content and is already very high on the first token for common content. This is evidenced also by the DreamSim score for each experiment, which is significantly lower for the common image experiments.

\section{Related Works}

\paragraph{\textbf{Privacy and Copyright Infringements}}
The intersection of privacy and copyright infringements in generative models has garnered significant attention. 
This approach assumes that to avoid copyright infringement, the output of a model shouldn't be too sensitive to any of its individual training samples. Bousquet et al. \cite{bousquet2020synthetic} suggest the use of differential privacy \cite{dwork2008differential} to stabilize the algorithm and avoid such sensitivity. Vyas et al. \cite{vyas2023provable} introduces a slightly less stringent notion (Near-Access Freenes) but relies on a similar benchmark, a \emph{safe model} that doesn't have access to the copyrighted data. Carlini et al. \cite{carlini2023extracting} and Haim et al. \cite{haim2022reconstructing} further explored this area by investigating the extraction of training data from models, highlighting the risks of memorization. \citet{elkin2023can}, however, investigated the gap between privacy and copyright infringement from the perspective of the law, and showed that requiring such notions of stability may be too strong, and are not always aligned with the original intention of the law. Closer to our approach, \cite{scheffler2022formalizing} suggests a framework to quantify originality by measuring the description length of a content with and without access to the allegedly copyrighted material. Our approach of textual inversion also looks for a succinct description of the content but, distinctively, our definition depends on the distribution of the data, and measures originality with respect to the whole data to be trained. This may lead to different outcomes, for example, when the allegedly copyrighted material contains a distinctive trait that is not necessarily original.

\paragraph{\textbf{Attribution in Generative Models}}
Attribution in generative models is a crucial area of research, focusing on identifying the sources of data that contribute to the generation of specific outputs. Park et al. introduced the TRAK method to address data attribution in large-scale models \cite{park2022}, and recently, Wang et al. \cite{wang2023evaluating} proposed a method for evaluating the attribution in Stable Diffusion models of data points in the generation process, which is closely related to assessing the originality of generated images. However, such a method requires full access and knowledge of the training set on which the model was trained.

\paragraph{\textbf{Generalization and Memorization}}
The interrelation between generalization and memorization is a key challenge for Machine Learning.
Classically, memorization and generalization are considered to be in tension. Ideal learning would seem to \emph{extract} relevant information but avoid memorizing irrelevant concepts. While limiting memorization does lead to generalization \cite{russo2019much, bassily2018learners, xu2017information, arora2018stronger}, recent studies suggest that memorization may be critical, and unavoidable in certain tasks \cite{feldman2020does, feldman2020neural, livni2024information}. Most recently, Attias et al. \cite{attias2024information} demonstrated how even in simple tasks such as mean estimation, memorization of the data is a prerequisite. On a practical level, Zhang et al. \cite{zhang2016understanding} explored the relationship between these two aspects, emphasizing their importance in the effectiveness of deep learning models.

\section{Discussion}
\label{sec:conclusion}

In this work, we introduced a novel approach to assess the originality of images with Text-to-Image (T2I) Generative Diffusion models, and have investigated its behavior in this aspect under a controlled environment. Our methodology leverages the concept of familiarity within the model's training data to quantify the originality of tested images. By employing textual inversion techniques, we demonstrated that the number of tokens required to represent and reconstruct an image serves as a measure of its originality, without requiring access to the training data, nor a specific prompt that potentially poses copyright complications. 

Our analysis confirmed that T2I models are capable of producing new original content, highlighting the importance of training models on diverse and comprehensive datasets. These findings also challenge the traditional view of avoiding memorization in models. Instead, we propose that models should familiarize themselves with a broad spectrum of data, respecting copyright constraints, to enhance their ability to generate new content.

In summary, our study offers a fresh perspective on evaluating originality in the context of generative models, which can inform copyright analysis and assist in delineating the legal protection afforded to such images more efficiently and accurately. By quantifying the familiarity of concepts to the model, we provide insights that align with legal definitions and can aid in addressing copyright eligibility, infringement, and licensing issues. In addition to law-related applications, our approach opens up new avenues for research in the intersection of generative models, originality assessment, and generative quality.

\paragraph{Limitations}
\label{sec:limitations}
One of the primary constraints for the method is the reliance on textual inversion, which may not capture all aspects of originality in complex images. Additionally, our method's effectiveness is contingent on the quality and diversity of the training data, which might not always be optimal. Furthermore, the correlation between token count and originality, although significant, may not be universally applicable across different model architectures or datasets. Future research should explore alternative measures of originality and test the robustness of our approach across a broader range of models and data, making it readily available for deployment. Finally, our work demonstrates that T2I models can be utilized to discriminate original and non-original work. That being said, an important motivation of our work is to assess originality of T2I content. Designing a framework that exploits generative model's ability to discriminate original content in order to audit genAI and safeguard content leads to several open problems and challenges which we leave to future work.

\section*{Impact Statement}
This paper presents work whose goal is to advance the field of Machine Learning by introducing a framework for quantifying originality in text-to-image generative diffusion models. The potential broader impact of this work includes the following:

\paragraph{Ethical Aspects}
Our research addresses the challenge of quantifying originality, which has significant implications for copyright laws and the protection of creative works. By providing a methodology to assess the originality of generated images, we aim to contribute to a fairer and more transparent use of generative models in creative industries. This could help mitigate legal disputes related to copyright infringement and ensure that the rights of original content creators are respected.

\paragraph{Future Societal Consequences}
The ability to quantify originality in generated images could enhance the deployment of generative models in various fields, including art, design, and entertainment, by fostering trust and accountability. It can also encourage the development of new creative tools that assist artists in generating unique content while respecting intellectual property rights.

While the primary goal of this work is to advance the field of Machine Learning, we believe that our contributions to the understanding of originality and creativity in generative models will have a positive societal impact by promoting ethical use and fostering innovation.

\paragraph{Acknowledgments} 
This research was funded in part by an ERC grant (FOG, 101116258), an ISF Grant (2188 $\backslash$ 20), and supported by a grant from The Center for AI and Data Science at Tel Aviv University (TAD). Views and opinions expressed are, however, those of the author(s) only and do not necessarily reflect those of the European Union or the European Research Council Executive Agency. Neither the European Union nor the granting authority can be held responsible for them. In addition, the research leading to these results was supported by TILabs Tel-Aviv University Innovation Labs and has been partially published by the Israeli Science Foundation 1337/22. We also thank illustrator Daniel Goldfarb for providing new original creations for this experiment \url{https://danielgoldfarbart.com}, Bruria Friedman and Ahava Azan for curating the common and original data sets.

\bibliography{icml2024}
\bibliographystyle{icml2024}
\clearpage
\appendix
\onecolumn

\clearpage
\section{Additional Results}
\label{sec:app_additional_results}

\begin{figure*}[h]
    \centering
\includegraphics[width=0.9\linewidth]{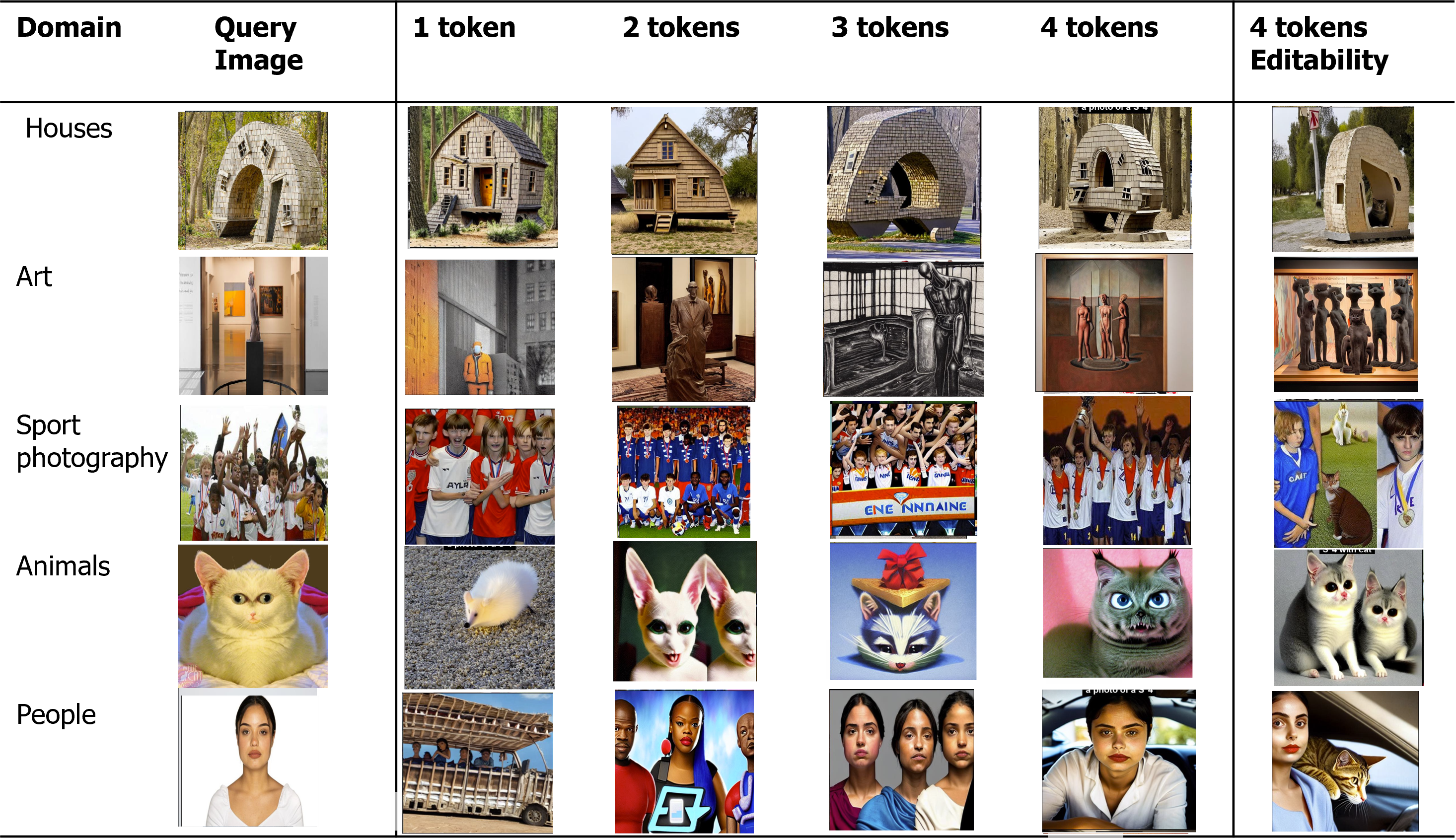}
    \vspace{-0.2cm}
    \caption{Addtional Qualitative results demonstrate the effectiveness of multi-token textual inversion in reconstructing original images across different domains, with more tokens enhancing the capture of additional details. On the right, we demonstrate the editability test using the prompt "Cat with \sm" for $m=4$.}
    \label{fig:sup_res_stable_original}
    \vspace{-0.3cm}
\end{figure*}
\begin{figure*}[h]
    
    \centering
\includegraphics[width=0.95\linewidth]{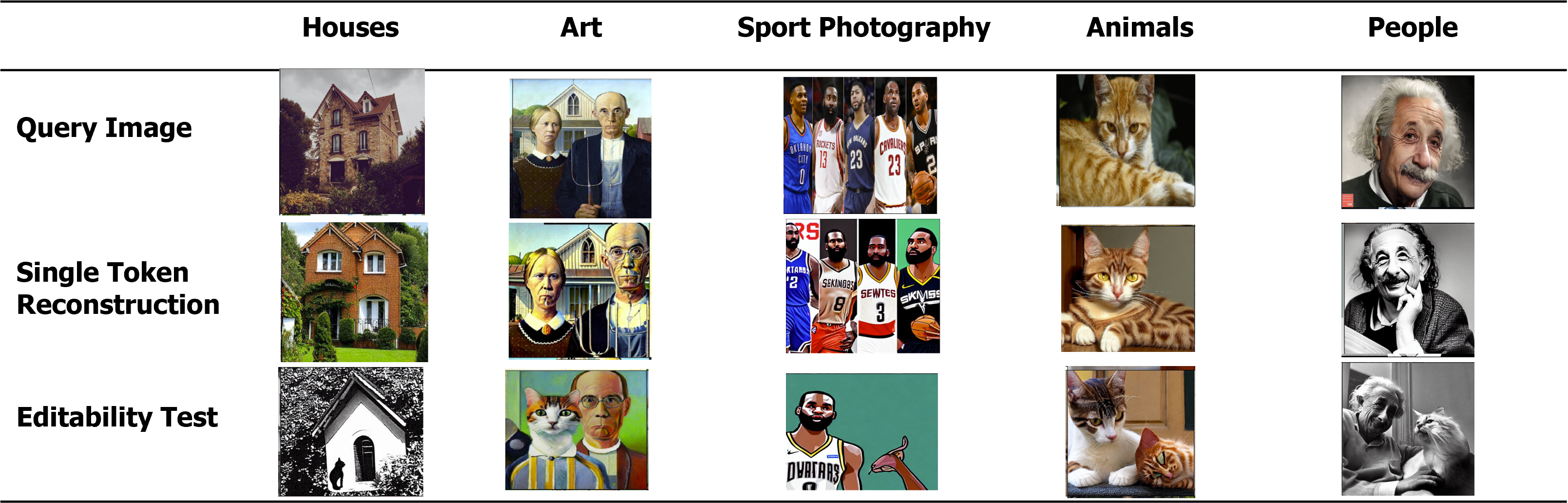}

    \caption{Additional qualitative results show that multi-token textual inversion can effectively reconstruct common images from various domains, with a single token often sufficing for high-quality reconstruction. The editability test in the last row is illustrated using the prompt "cat with \sm".}
    \label{fig:sup_res_stable_common}

\end{figure*}

We include additional results for both Common and Original categories in each domain. Results for Houses, Sports Photography, Animals, Art, and People from the original set are shown in~\cref{fig:sup_res_stable_original}, and from the common set in~\cref{fig:sup_res_stable_common}.

\clearpage
\section{Synthetic Framework}
\label{sec:appx_syn_framework}
\subsection{Implementation Details}
\label{sec:apx_syn_impl}

\paragraph{\textbf{Datasets}} All images feature a white background and a number of geometric shapes in the foreground (see~\cref{fig:synthetic_gen_diag}(ii)).  Each \textit{element} in the image is defined by four features, and the entire set is the cross-product of all features. The shapes are independently and uniformly located across the image. The image's default textual description is in the format \textit{"big red full circle and small empty blue square"} \footnote{Based on empirical experiments, we have found this more effective than counting or grouping elements in the prompt. Evidence for misalignment between the prompt and produced images has been shown in various studies as well \cite{chefer2023attend, wang2023decompose}}.

Utilizing this framework, we generate datasets comprising 100K images each. In every dataset, 10\% of the images are empty, while the rest contain a variable number of elements uniformly distributed within the range $[1,n]$, where $n$ can be any natural number. In the following sections, we choose either $n=4$ or $n=6$. 

\paragraph{\textbf{Models}} We train a separate Stable Diffusion model for each dataset, so that the only visual data seen by the resulting model is the training dataset itself.  Key decisions included (i) pretraining a VAE (Encoder-Decoder) and a UNet (noise-cleaner) from scratch, and (ii) employing BERT as the Text-Encoder, chosen over CLIP to avoid the broader visual context implications associated with CLIP training.

\paragraph{\textbf{Evaluation}} In order to facilitate automated and large-scale analysis of generations, which is essential for the purposes outlined in \S\ref{sec:analysis}, we fine-tuned a YOLOv8 model on the synthetic datasets.
This approach effectively addresses common issues, such as overlapping or slightly deformed elements, providing a confidence measure for each detection. We set the confidence threshold at 0.9, aligning with the requisite quality of the generated elements. 

\subsection{Generalization Experiments Additional Details}
\paragraph{\textbf{Generalization target}} In assessing generalization, we leave out specific elements from the training process, and ask for their generation after training. While these elements have not been witnessed by the model, their properties have. For example, if a blue circle is omitted from training, the model still witnesses circles and the color blue, only not in conjunction (Fig.~\ref{fig:synthetic_gen_diag}(i), bottom). 
The degree of generalization can be evaluated by the frequency of occurrence of the missing element within the generated set, normalized by the total number of generated elements. We repeat this experiment with a prompt asking for the specific missing element, and unconditionally with the empty prompt.

\paragraph{\textbf{Text conditioning}} In evaluating each trained model, we assess the occurrence frequency of the generated missing element across two sets, each comprising 1024 generated images. Initially, we generate images using an empty prompt (i.e., " "), thereby sampling from the unconditioned distribution represented by the model. Subsequently, we generate images with a prompt precisely describing the missing element (e.g., "blue circle"), thereby sampling from the model's text-conditioned distribution. It is natural to anticipate that employing a specific textual prompt will increase the frequency of the generated missing element.

\paragraph{\textbf{Training data diversity}} As discussed in \S\ref{sec:exp}, each element within our dataset encompasses values of four dimensions: Size, Color, Texture, and Shape Type.  Our dataset's diversity spectrum ranges from the least diverse, characterized by a span of two shape types (square and circle) and two colors (red and blue) (Fig.~\ref{fig:synthetic_gen_diag} bottom left), to the most diverse, which encompasses four dimensions - five shape types (square, circle, triangle, hexagon, and star), three colors (red, green, and blue), two sizes (big and small), and two textures (full and empty), thereby resulting in 60 unique elements (Fig.~\ref{fig:synthetic_gen_diag} top left). Consistent with prior research \cite{zhao2018bias}, we anticipate a positive correlation between diversity and generalization. 

\paragraph{\textbf{Addressing Bias}} To mitigate potential bias arising from the model's inclination towards certain values within our element subspace, we enforce symmetry by averaging over a larger number of experiments, each differing in its spanning set and missing elements. For instance, the leftmost data points in Fig.~\ref{fig:analysis_results} represent the averaged results of four identical 4-element experiments conducted sequentially with (1) big full elements, (2) big empty elements, (3) small full elements, and (4) small empty elements.

\section{In-Distribution Assessment in the Synthetic Setting Ablation Study}
\label{sec:apx_syn_indistribution_ablation}
In Section 4, we outline simplified criteria for in-distribution testing within a synthetic setting vs a real-world setting, where we employ the concept of edibility as a metric. In the synthetic domain, a crucial test for the model's effectiveness lies in its ability to generate images without merely copying the spatial placement of elements from the query image. To test this, we provide the following ablation study, where we fixed the location of elements in common images during the training phase. The rationale behind this methodology was to challenge the notion that the model's generation of elements in varied locations might still be indicative of overfitting and to ensure that the model stays within the intended distribution bounds.

Given that our model operates on patches, it could be suggested that if the model recreates identical elements in different locations, it might not be exhibiting true understanding but rather a form of overfitting. To address this, we trained the model on images with a single fixed location, hypothesizing that if the model were able to replicate these elements in the same fixed location, it would demonstrate an awareness of element locations beyond mere memorization.

The results of the ablation study supported our premise: the elements from the common images consistently appeared in the same spatial positioning as in the query image, providing evidence of the model's spatial awareness. This finding is vital as it suggests that the model's generation of elements in different locations is not an artifact of overfitting but rather an indication of its genuine understanding of elements.

\begin{figure}[h]
    \centering
    \vspace{-0.1cm}
    \includegraphics[width=0.8\linewidth]{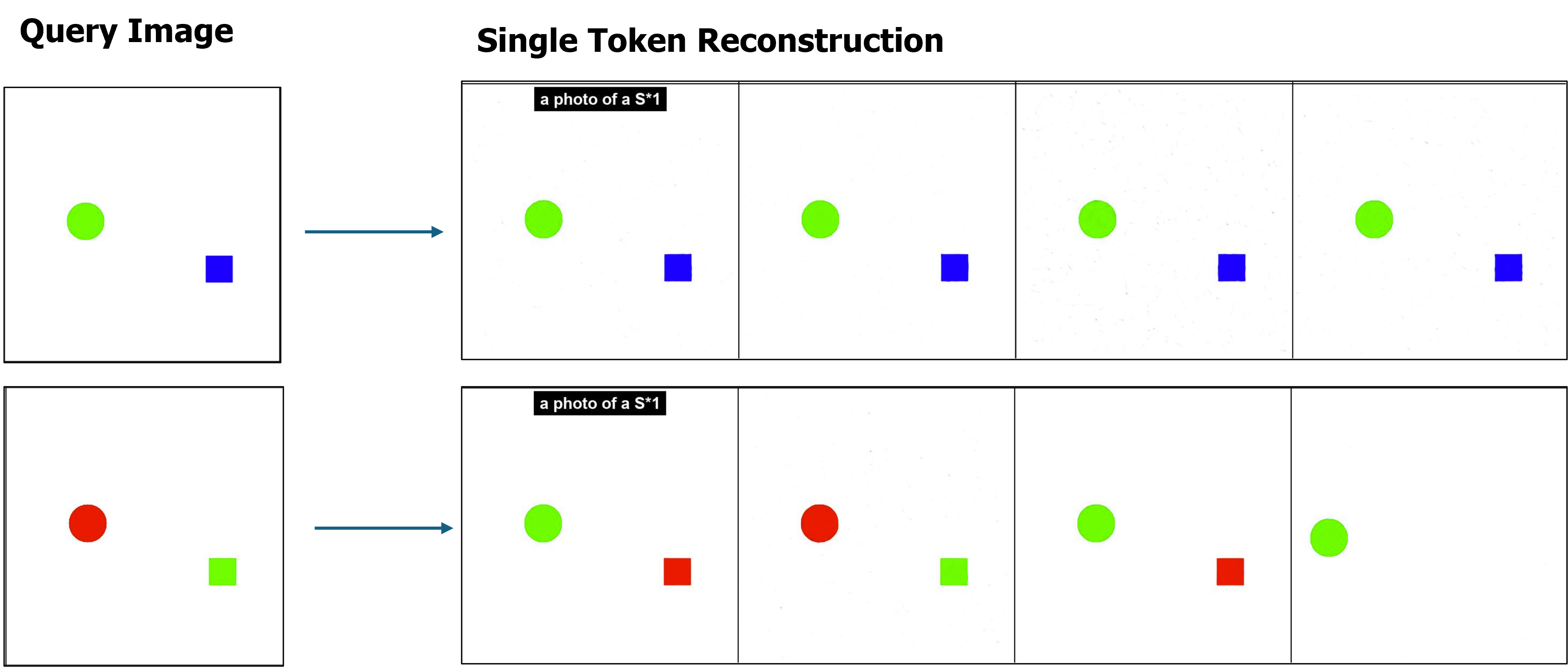} 
    \vspace{-0.2cm}
    \caption{Qualitative Illustrations of the Ablation Study. The original query images on the left and their corresponding single-token reconstructions on the right, generated using four distinct random seeds. These examples serve to validate the model's capability to comprehend and maintain the fixed spatial locations of elements as observed in the common images during training, demonstrating the validity 
 of our in-distribution test in the synthetic setting.}
    \label{fig:sup_syn_ablation}
    \vspace{-0.4cm}
\end{figure}

Qualitative illustrations from this study are presented in Fig.~\ref{fig:sup_syn_ablation}, with the original query images on the left and the single-token reconstructions generated using four distinct random seeds on the right.

\clearpage
\section{Synthetic Images Quality Analysis}
\begin{figure}[h]
    \centering
    \includegraphics[width=\linewidth]{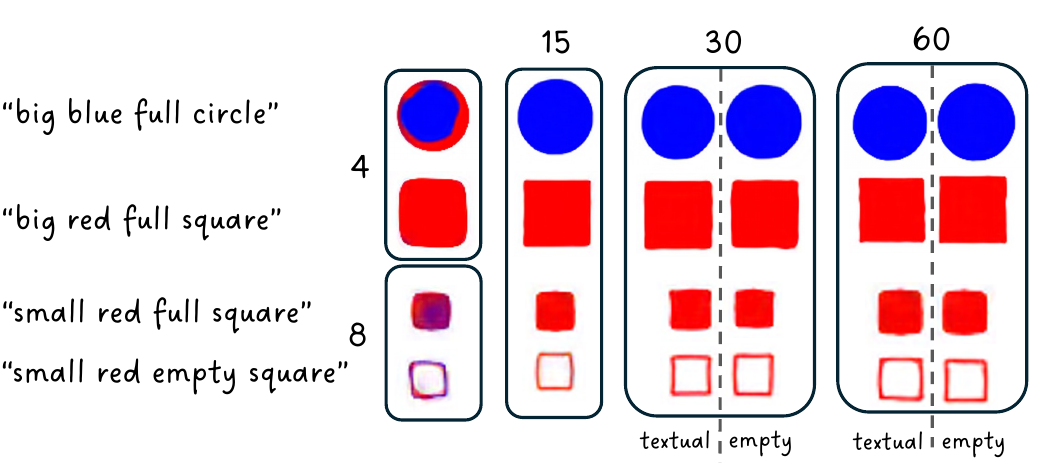} 
    \caption{Detailed set of generated samples showcasing the correlation between training data diversity and generated images quality. \textbf{Rows:} All elements in a row are missing elements of the same class, when appearing in generated images. \textbf{Blocks:} All elements in a block where generated from models trained on data of the same diversity scale, ranging from 4 elements to 60. \textbf{Sub-blocks:} In the 30 and 60 blocks, the right-hand side sub-blocks represent the results of empty prompts, and the left-hand side sub-blocks represent the results of ad-hoc textual prompts.}
    \label{fig:quality_vs_diversity}
\end{figure}

As illustrated in Fig.~\ref{fig:quality_vs_diversity}, we observed an expected relationship between training data diversity and generated image quality. The quality of generated elements not present in the training set improves with greater diversity. Additionally, we found that the generation quality was not impacted by the type of conditioning, particularly in more diverse cases where unconditioned generation led to the creation of missing elements.

\begin{figure*}[t]
    \centering
    \vspace{-0.1cm}
    \includegraphics[width=.99\linewidth]{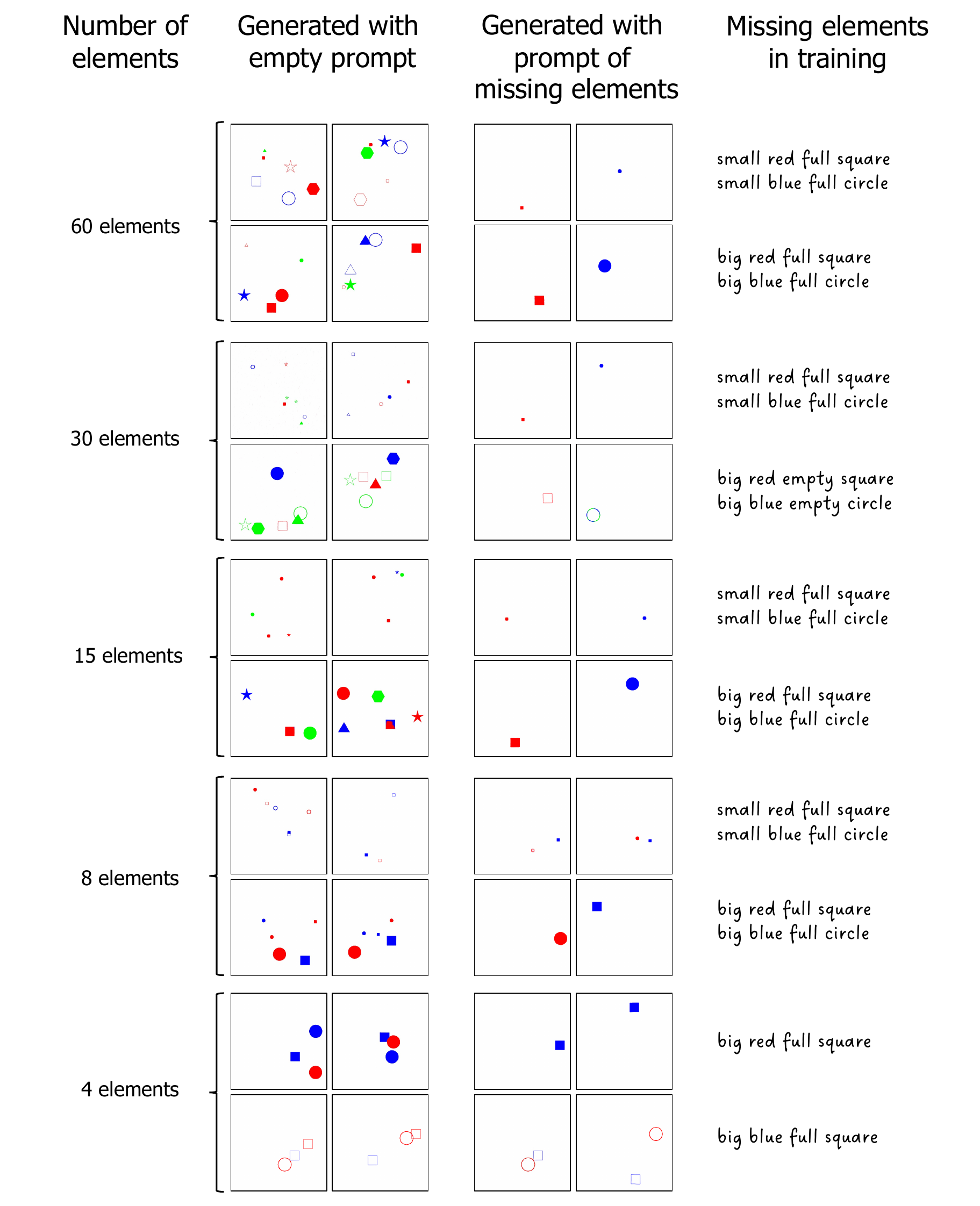} 
    
    \vspace{-0.2cm}
    \caption{A representative sample of generations made by the synthetic models. Each row contains images generated by a different model. The models differ by the number of different element in their training set, and the specific elements left out. The left columns contain generations with an empty prompt, while the right columns contain images generated with a prompt describing one of the elements missing in training.}
    \label{fig:syn_generated_samples}
    \vspace{-0.4cm}
\end{figure*}

In addition, we provide sample generations from each of the experiments in this study 
in~\cref{fig:syn_generated_samples}.

\clearpage
\section{Quantifying Originality using Textual Inversion: Additional Implementation Details}
\label{sec:apx_text_inversion_impl}

This section provides detailed information on the implementation of the models described in Section 4.

\subsection{Synthetic Experiments}
\paragraph{\textbf{Stable Diffusion Pre-Training Details}} Our T2I model training involved two stages. First, we trained a VAE for 8 epochs with an effective batch size of 32 and a learning rate of $10^{-5}$. Next, we trained a UNet for 15 epochs using the trained VAE and a pre-trained BERT model, with an effective batch size of 64 and a learning rate of $6.4 \times 10^{-5}$.

We evaluated our method's originality assessment in a controlled environment by synthesizing a custom dataset with specified features: Type: [circle, square, triangle, hexagon, star], Color: [red, green, blue], Size: [big], Texture: [full]. The dataset was structured as follows:
\begin{itemize}
\item \textbf{Common:} 30\% of the images contain the pair (circle, square) in varying colors.
\item \textbf{Rare:} 0.1\% contain the pair (circle, triangle) in varying colors.
\item \textbf{Unseen:} The pair (square, triangle) does not appear in any images.
\end{itemize}
The remaining images may contain up to 4 elements, with no more than one from the set [circle, square, triangle]. We consider the frequent pair (circle, square) as generic, with multiple variations in each image's element positions. Our method quantifies originality based on this stable diffusion model, using a total of 100K instances for training.

\paragraph{\textbf{Multi-Tokens Textual Inversion Training Details}}
In the synthetic setup, we trained our multi-token textual inversion variant with token lengths ranging from 1 to 5. The training used a batch size of 20, a learning rate of 0.0005, and 2000 steps. We found that 50 denoising inference steps produced cleaner results.

\subsection{Pretrained Stable Diffusion Experiments}
\paragraph{\textbf{Reconstruction Measurement}}
We measure the similarity between original and reconstructed images using the DreamSim distance \cite{fu2023dreamsim}. DreamSim is built upon an ensemble of different models; for our use case, we used the DreamSim distance, which includes all models. At evaluation time we generate for each image 20 images using the prompt \texttt{"a photo of \sm"} and average the dreamsim score.  

\paragraph{\textbf{Training Prompt Templates}}
Similar to the original Textual Inversion method, we used object text templates for all experiments except the art domain, following the approach in [reference]. For the art domain, we used a custom list generated by GPT4 and manually curated. The full list of text templates includes:
\begin{itemize}
\item "a detailed image of the artwork titled \sm"
\item "a high-resolution photo of the artwork \sm"
\item "a close-up view of the artwork known as \sm"
\item "a digital representation of the art piece \sm"
\item "the famous artwork \sm"
\item "a full view of the art piece titled \sm"
\item "an artistic interpretation of \sm"
\item "a gallery display of the artwork \sm"
\item "a photographic capture of the art \sm"
\item "the artwork \sm in full detail"
\item "a visual study of the artwork \sm"
\item "the complete artwork known as \sm"
\item "an exhibition view of \sm"
\item "a curated image of the artwork \sm"
\item "a detailed scan of \sm"
\item "an artistic rendering of \sm"
\item "a high-quality image of the artwork \sm"
\item "the full artwork titled \sm"
\item "a museum display of \sm"
\item "an archival photograph of the artwork \sm"
\end{itemize}

\section{DreamSim Distance Metric}
\label{sec:apx_dreamsim}

In section~\cref{sec:method}, we describe our method for measuring originality, which includes measuring the distance between the query image and the reconstructed image.
DreamSim is an advanced perceptual image similarity metric with STOA performances that offers a more comprehensive and human-aligned approach to evaluating image similarity compared to traditional methods like the Fréchet Inception Distance (FID).

DreamSim is designed to bridge the gap between low-level image metrics (such as LPIPS, PSNR, and SSIM) and high-level semantic judgments (such as those made by models like CLIP). Traditional metrics often fall short in capturing mid-level differences in image layout, object pose, and semantic content, which are crucial for aligning with human visual perception.

DreamSim leverages embeddings from several pre-trained models, including CLIP \cite{radford2021learning}, OpenCLIP \cite{ilharco_gabriel_2021_5143773}, and DINO \cite{caron2021emerging}. These embeddings are fine-tuned using human perceptual judgments on a dataset of synthetic images created by text-to-image models. The fine-tuning process involves learning from around 20,000 image triplets, where human annotators have determined which images are more similar.

The formulation of DreamSim can be summarized as follows:
\begin{itemize}
    \item \textit{Concatenation and Fine-tuning:}
   The embeddings are concatenated and fine-tuned on human perceptual judgments:
   $
   E_{\text{concat}} = \text{concat}(E_{\text{CLIP}}, E_{\text{OpenCLIP}}, E_{\text{DINO}})
   $, 
   $
   E_{\text{DreamSim}} = \text{fine-tune}(E_{\text{concat}}, \text{human\_judgments})
   $
   Where, $( E_{\text{CLIP}}, E_{\text{OpenCLIP}}, E_{\text{DINO}} )$ are the embedding functions of the respective models.
   \item \textit{Cosine Similarity:}
   The perceptual distance \( D \) between two images \( I_1 \) and \( I_2 \) is computed as the cosine distance between their embeddings:
   \[
   D(I_1, I_2) = 1 - \frac{E_{\text{DreamSim}}(I_1) \cdot E_{\text{DreamSim}}(I_2)}{\|E_{\text{DreamSim}}(I_1)\| \|E_{\text{DreamSim}}(I_2)\|}
   \]
\end{itemize}

\paragraph{Advantages over Traditional Metrics}
\begin{itemize}
    \item{\textit{Human Alignment}: DreamSim is trained on human judgments, making its similarity assessments more aligned with how humans perceive visual similarity.}
    \item{\textit{Comprehensive Feature Capture}: By using embeddings from multiple models, DreamSim captures a wide range of visual features, from low-level textures to high-level semantic content.}
    \item{\textit{Generalization:} Despite being trained on synthetic data, DreamSim generalizes well to real images, making it versatile for various applications, including image retrieval and reconstruction tasks.}
\end{itemize}

\end{document}